\documentclass[journal,comsoc]{IEEEtran}

\usepackage[T1]{fontenc}% optional T1 font encoding

\usepackage{graphicx}
\usepackage{amsmath}
\usepackage{amssymb}
\usepackage{color}
\usepackage[misc]{ifsym}
\usepackage{subfigure}
\usepackage{array}
\usepackage{enumerate}
\usepackage{multirow}
\usepackage{verbatim}
\usepackage{cite}
\usepackage{url}
\usepackage{algorithm}
\usepackage{algorithmic}
\usepackage{amsmath}
\usepackage{verbatim}
\usepackage{color}
\usepackage[table]{xcolor}
\ifCLASSINFOpdf

\else

\fi

\usepackage{amsmath}

\usepackage[cmintegrals]{newtxmath}
\usepackage{algorithm}
\usepackage{algorithmic}
\usepackage{amsmath}
\usepackage{verbatim}

\begin{document}

\title{A Comprehensive Review of Agricultural Parcel and Boundary Delineation from Remote Sensing Images: Recent Progress and Future Perspectives}

\author{Juepeng Zheng, ~\IEEEmembership{Member,~IEEE}, 
        Zi Ye, 
        Yibin Wen, 
        Jianxi Huang, ~\IEEEmembership{Senior Member,~IEEE},   
        Zhiwei Zhang, \\
        Qingmei Li, 
        Qiong Hu, 
        Baodong Xu, ~\IEEEmembership{Member,~IEEE},  
        Lingyuan Zhao, 
        Haohuan Fu, ~\IEEEmembership{Senior Member,~IEEE} 
        %,~\IEEEmembership{Life~Fellow,~IEEE}% <-this % stops a space
\thanks{Manuscript received. This work was supported by the National Natural Science Foundation of China (Grant No. 42401415 and T2125006); Shenzhen Science and Technology Program under Grant KCXFZ20240903093759004 and Grant KJZD20230923115106012; Fundamental Research Funds for the Central Universities, Sun Yat-sen University (Project No. 24xkjc002) and Jiangsu Innovation Capacity Building Program (Project No. BM2022028). \textit{(Corresponding authors: Jianxi Huang and Haohuan Fu.)}}

\thanks{Juepeng Zheng is with the School of Artificial Intelligence, Sun Yat-Sen University, Zhuhai, China, and National Supercomputing Center in Shenzhen, Shenzhen, China (Email: zhengjp8@mail.sysu.edu.cn).}% <-this % stops a space

\thanks{Zi Ye, Yibin Wen and Zhiwei Zhang are with the School of Artificial Intelligence, Sun Yat-Sen University, Zhuhai, China (Email: { , zhangzhw65}@mail2.sysu.edu.cn).}% <-this % stops a space

\thanks{Jianxi Huang is with the Faculty of Geosciences and Engineering, Southwest Jiaotong University, Chengdu, China, the College of Land Science and Technology, China Agricultural University, Beijing, China, and Key Laboratory of Remote Sensing for Agri-Hazards, Ministry of Agriculture and Rural Affairs, Beijing, China (Email: jxhuang@cau.edu.cn).}

\thanks{Qingmei Li is with the Tsinghua Shenzhen International Graduate School, Tsinghua University, Shenzhen, China (Email: qingmeili1997@gmail.com).}

\thanks{Qiong Hu is with the College of Urban and Environmental Sciences, Central China Normal University, Wuhan, China (Email: huqiong@ccnu.edu.cn).}

\thanks{Baodong Xu is with the Macro Agriculture Research Institute, College of Resources and Environment, Huazhong Agricultural University, Wuhan, China (Email: xubaodong@mail.hzau.edu.cn).}

\thanks{Lingyuan Zhao is with the HuanTian Wisdom Technology Co., Ltd., Meishan, Sichuan (Email: zhaoly@htwisdom.cn).}

\thanks{Haohuan Fu is with Tsinghua Shenzhen International Graduate School, Tsinghua University, Shenzhen, China, the Ministry of Education Key Laboratory for Earth System Modeling, Department of Earth System Science, Tsinghua University, Beijing, China, National Supercomputing Center in Shenzhen, Shenzhen, China (Email: haohuan@tsinghua.edu.cn).}

%\thanks{Le Yu is with the Ministry of Education Key Laboratory for Earth System Modeling, Department of Earth System Science, Tsinghua University, Beijing, China (Email: leyu@tsinghua.edu.cn).}

}

\markboth{Preprint Version}%Submit to IEEE Geoscience and Remote Sensing Magazine
{Shell \MakeLowercase{\textit{Zheng et al.}}: Bare Demo of IEEEtran.cls for IEEE Communications Society Journals}

\maketitle

\begin{abstract}
Powered by advances in multiple remote sensing sensors, the production of high spatial resolution images provides great potential to achieve cost-efficient and high-accuracy agricultural inventory and analysis in an automated way. 
Lots of studies that aim at providing an inventory of the level of each agricultural parcel have generated many methods for Agricultural Parcel and Boundary Delineation (APBD). 
This review covers APBD methods for detecting and delineating agricultural parcels and systematically reviews the past and present of APBD-related research applied to remote sensing images. 
With the goal to provide a clear knowledge map of existing APBD efforts, we conduct a comprehensive review of recent APBD papers to build a meta-data analysis, including the algorithm, the study site, the crop type, the sensor type, the evaluation method, etc. 
We categorize the methods into three classes: (1) traditional image processing methods (including pixel-based, edge-based and region-based); (2) traditional machine learning methods (such as random forest, decision tree); and (3) deep learning-based methods. 
With deep learning-oriented approaches contributing to a majority, we further discuss deep learning-based methods like semantic segmentation-based, object detection-based and Transformer-based methods. 
In addition, we discuss five APBD-related issues to further comprehend the APBD domain using remote sensing data, such as multi-sensor data in APBD task, comparisons between single-task learning and multi-task learning in the APBD domain, comparisons among different algorithms and different APBD tasks, etc. 
Finally, this review proposes some APBD-related applications and a few exciting prospects and potential hot topics in future APBD research. We hope this review help researchers who involved in APBD domain to keep track of its development and tendency.
\end{abstract}


\IEEEpeerreviewmaketitle

\section{Introduction}
\label{intro}

\IEEEPARstart{A}{gricultural} parcels represent the fundamental spatial units for agricultural activities, and their precise delineation is crucial for efficient resource management and a wide range of critical applications, including crop mapping \cite{alami2023crop}, yield estimation \cite{xiao2025progress}, and sustainable agricultural planning \cite{jones2017toward}. As the foundational layer for most geospatial agricultural analyses, the accuracy and quality of parcel boundary extraction directly influence downstream processes in precision agriculture \cite{weiss2020remote}, land-use monitoring \cite{wu2023challenges}, and policy implementation \cite{pe2019greener}. 
Traditionally, field boundaries have been delineated manually or derived through rule-based image processing techniques that rely on spectral thresholds or vegetation indices. However, these methods often fall short in complex agricultural landscapes characterized by irregular field shapes, heterogeneous crop conditions, and indistinct or occluded boundaries resulting from natural variability or farming practices. Moreover, manual annotation is inherently time-consuming, labor-intensive, and unsuitable for large-scale or frequent updates. 
With the advent of high-resolution commercial satellites and rapid advancements in computational techniques, \underline{\textbf{A}}gricultural \underline{\textbf{P}}arcel and \underline{\textbf{B}}oundary \underline{\textbf{D}}elineation (\textbf{APBD}) has entered a new era. Modern approaches have evolved from classical machine learning algorithms, such as Random Forests (RFs) and Support Vector Machine (SVM), to deep learning frameworks, including Convolutional Neural Networks (CNNs) \cite{lecun2015deep} and U-Net \cite{ronneberger2015u}. More recently, the emergence of foundation models (such as Transformer architectures \cite{han2022survey}, the Segment Anything Model (SAM) \cite{kirillov2023segment}, and Vision-Language Models (VLMs) \cite{zhang2024vision}) has further expanded the capabilities of APBD. These models, leveraging massive datasets and self-supervised learning, offer strong generalization ability, high adaptability, and improved performance across diverse agricultural scenarios when applied to high-resolution remote sensing imagery.

\begin{figure*}[t]
    \centering
    \includegraphics[width=1.0\linewidth]{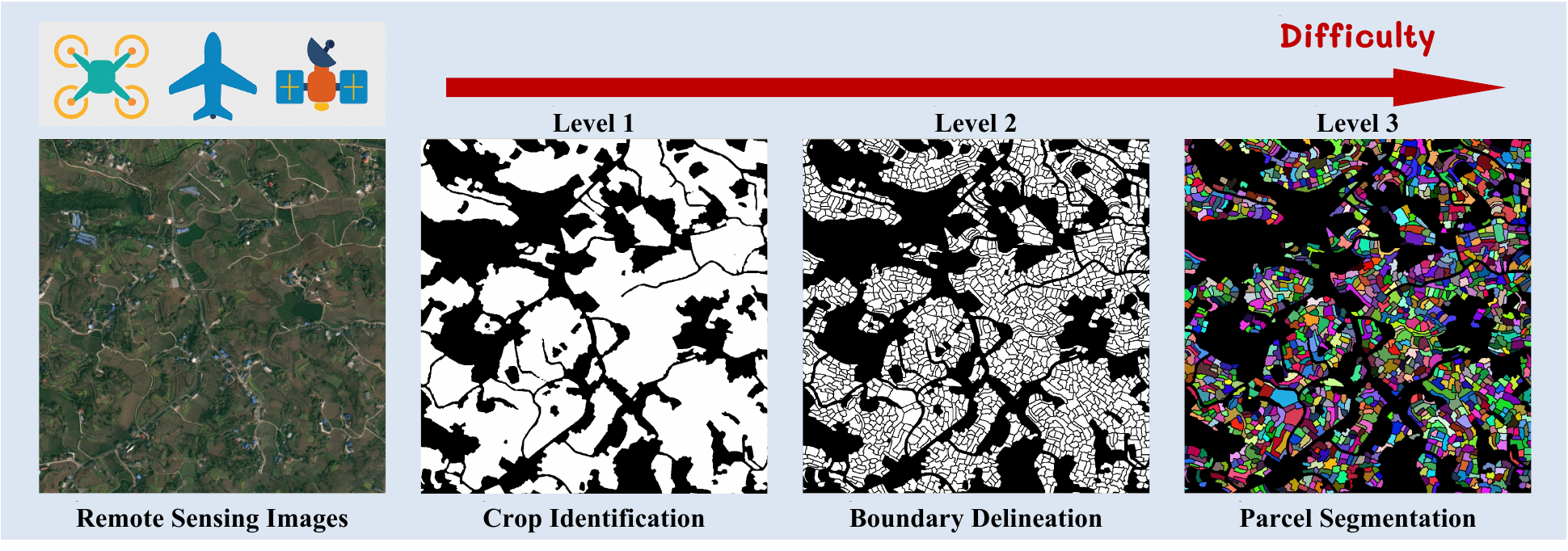}
    \caption{Three hierarchical levels for APBD in this review, including Cropland Identification (CI), Boundary Delineation (BD), and Parcel Segmentation (PS).}
    \label{fig:three}
\end{figure*}

\begin{table*}[t]
    \centering
    \caption{\label{tab:SOTAreview} {Summary of existing APBD related reviews. TIP, TML and DL denote traditional image processing, traditional machine learning and deep learning-based APBD methods. CI, BD and PS denote three different levels of APBD task, i.e., Cropland Identification, Boundary Delineation and Parcel Segmentation. %Notably, $^*$ denotes this paper is a comparative study instead of a review paper and it only reviews 9 deep learning-based methods.
    } }
    %\resizebox{\textwidth}{!}
    {
    \begin{tabular}{ccccccccc}
    \hline
       \multirow{2}*{Review Articles} &  \multicolumn{3}{c}{Reviewed APBD methods} &  \multicolumn{3}{c}{Reviewed APBD tasks} & \multirow{2}*{Datasets} &  \multirow{2}*{Applications} \\ %\multicolumn{2}{c}{Reviewd Data} & 
       &  TIP & TML & DL &  CI & BD & PS   \\\hline %Passive RS & Active RS &
       Nagraj \textit{et al.} \cite{nagraj2016crop} &  $\checkmark$ & $\checkmark$ & $\times$ &  $\checkmark$ &  $\times$ & $\times$ & $\times$ & $\times$ \\%$\times$ & $\checkmark$ &
       Joshi \textit{et al.} \cite{joshi2023remote} & $\times$ & $\times$ & $\checkmark$ & $\checkmark$ &  $\times$ & $\times$ & $\times$ & $\checkmark$  \\%$\checkmark$ & $\times$ & 
       Alami \textit{et al.} \cite{alami2023crop} & $\times$ & $\checkmark$ & $\checkmark$  & $\checkmark$ & $\times$ & $\times$ & $\times$ & $\times$ \\%$\times$ & $\times$ & 
       Wang \textit{et al.} \cite{wang2023survey} & $\checkmark$ & $\checkmark$ & $\times$ &  $\times$ & $\checkmark$ & $\times$  & $\times$ & $\times$ \\%$\checkmark$ & $\times$ &
       % Wu \textit{et al.} \cite{wu2023challenges} \\
       Omia \textit{et al.} \cite{omia2023remote} & $\checkmark$ & $\checkmark$ & $\checkmark$ &  $\checkmark$ &  $\times$ & $\times$ & $\times$ & $\times$   \\%$\checkmark$ & $\checkmark$ &
       Xu \textit{et al.} \cite{xu2024deep} & $\times$ & $\times$ & $\checkmark$ &  $\times$ & $\checkmark$ & $\times$ & $\checkmark$ & $\times$ \\ \hline %$\checkmark$ & $\checkmark$ &
       %Hadir \textit{et al.}$^*$ \cite{hadir2025comparative} & $\times$ & $\times$ & $\checkmark$ &  $\checkmark$ &  $\checkmark$ & $\checkmark$ & $\checkmark$ & $\times$   \\ \hline %$\checkmark$ & $\times$ &
        \rowcolor{red!10} Ours & $\checkmark$ & $\checkmark$ & $\checkmark$ &  $\checkmark$ &  $\checkmark$ & $\checkmark$ & $\checkmark$ & $\checkmark$ \\  \hline %$\checkmark$ & $\checkmark$ &
    \end{tabular}
    }
     %    \begin{tablenotes}
	    % \footnotesize 
       
     %    \item[1] 
        
     %    \end{tablenotes}
    
\end{table*}

A wide range of review articles have extensively explored agricultural remote sensing applications \cite{weiss2020remote,you2022mapping,liang2024advancements,zhang2025review,zheng2025review}, covering core tasks such as crop mapping \cite{joshi2023remote,alami2023crop}, disease detection \cite{berger2022multi}, growth monitoring \cite{gao2021mapping,wu2023challenges,omia2023remote}, and yield estimation \cite{xiao2025progress}, among others. Some surveys focus on specific crop types, including paddy rice \cite{dong2016evolution,zhao2021mapping,fang2024comprehensive}, maize \cite{chivasa2017application}, and wheat \cite{manafifard2024comprehensive}. 
However, relatively few reviews have examined the APBD in depth, particularly in relation to relevant datasets and deep learning-based methodologies. To address this gap, we summarize existing literature in Table \ref{tab:SOTAreview}, highlighting distinctions in methods, topics, tasks, and data sources. For example, Wang \textit{et al.} \cite{wang2023survey} systematically analyze the farmland boundary extraction process, associated detection algorithms, and key influencing factors. Xu \textit{et al.} \cite{xu2024deep} combine bibliometric and content analysis to provide a comprehensive overview of deep learning applications in APBD. Hadir \textit{et al.} \cite{hadir2025comparative} focus specifically on the evaluation and comparative analysis of deep learning models for agricultural parcel delineation. By contrast, other surveys address only crop mapping (see Crop Identification (Level 1) in Fig. \ref{fig:three}), without considering agricultural parcel boundary delineation \cite{nagraj2016crop,joshi2023remote,alami2023crop,omia2023remote}.

According to Table \ref{tab:SOTAreview}, most existing APBD reviews are limited in scope and focus mainly on crop identification, overlooking boundary delineation and recent advances driven by deep learning and foundation models. 
First, most existing reviews retrieved and screened contain fewer than 100 relevant publications, reflecting a limited coverage of APBD and its associated tasks and applications, including crop classification, yield estimation, stress detection, and monitoring. 
Second, many of these reviews concentrate primarily on crop identification, often neglecting the equally critical task of boundary delineation. In this review, we conceptualize APBD as comprising three hierarchical levels: Cropland Identification (\textbf{CI}), Boundary Delineation (\textbf{BD}), and Parcel Segmentation (\textbf{PS}) (see Fig. \ref{fig:three} for details). \textbf{CI}, the simplest task, distinguishes cropland from background (a two-class classification problem). \textbf{BD} focuses on precisely delineating parcel boundaries, whereas \textbf{PS} is the most challenging, requiring the segmentation of each parcel instance individually. 
Furthermore, the current body of literature does not fully capture the breadth of recent advances in APBD, especially those driven by state-of-the-art deep learning algorithms and foundation models. These emerging techniques offer substantial improvements in accuracy, robustness, and scalability, yet remain insufficiently addressed in existing surveys. 

% First, most existing reviews retrieved and screened include a limited number of relevant publications (fewer than 100), indicating a lack of comprehensive and in-depth coverage of APBD and its associated tasks and applications, such as crop classification, yield estimation, stress detection, and monitoring. 
% Second, many of these reviews focus primarily on crop identification while overlooking the crucial task of boundary delineation. In this review, we categorize APBD into three hierarchical levels: Cropland Identification (CI), Boundary Delineation (BD), and Parcel Segmentation (PS) (see Fig. \ref{fig:three} for details). 
% Moreover, current literature fails to reflect the full spectrum of recent progress in APBD, particularly the advancements enabled by modern deep learning algorithms and foundation models. These emerging methods offer enhanced accuracy and scalability, yet remain underexplored in existing surveys.

% Therefore, it is essential to summarize the overall trends in APBD-related research over the past decade, providing readers with a comprehensive understanding of the historical development, current status, and future directions of this field. The main contributions of this review are as follows:
Therefore, it is essential to synthesize the overall trends in APBD research over the past decade, offering readers a comprehensive perspective on its historical evolution, current status, and future directions. The key contributions of this review are summarized as follows:

% \begin{enumerate}
%     \item We present a systematic and in-depth review of Agricultural Parcel and Boundary Delineation (APBD), including a meta-analysis of the literature, detailed comparison of methodologies, discussion of practical applications, and exploration of future research directions. To the best of our knowledge, this is the first comprehensive review specifically focused on APBD in the past decade. 
%     \item Cater to the rapid progress in computer science and its usage in APBD, we discuss the advantages and disadvantages of all kinds of existing APBD approaches from three aspects: traditional image processing approaches, traditional machine learning approaches, and deep learning approaches. We conduct comparisons between general deep learning models and their applications.
%     \item We conduct in-depth discussions on the multi-sensor data in APBD, dataset construction in APBD, comparison among different APBD algorithms, and analyze the criteria of choosing the proper methods. Also, we list extensive APBD-related applications and tasks, and envision promising future works in the APBD domain. We emphasize that remote sensing data will continue to be a key driver of future APBD-related studies.  
% \end{enumerate}

\begin{enumerate}
\item We provide the first systematic and in-depth review dedicated to Agricultural Parcel and Boundary Delineation (APBD) in the past decade. Our work includes a meta-analysis of existing literature, a detailed comparison of methodologies, an overview of practical applications, and an exploration of future research opportunities.
\item In response to the rapid advances in computer science and their growing adoption in APBD, we analyze the strengths and limitations of existing approaches from three perspectives: traditional image processing, traditional machine learning, and deep learning. We further compare general-purpose deep learning models with their tailored applications in APBD.
\item We conduct an in-depth discussion of multi-sensor data utilization in APBD, dataset construction, algorithmic comparisons, and criteria for selecting appropriate methods. We also provide an extensive overview of APBD-related applications and tasks, and outline promising research directions. We highlight that remote sensing data will remain a critical driver of APBD innovation in the foreseeable future.
\end{enumerate}

The rest content of this review is organized as follows. We present the meta-analysis of related literature in Sec. \ref{sec:meta}. Following that, we conduct a thorough review of the methodology of APBD in Sec. \ref{sec:method} and the assessment (such as public dataset and different evaluation metrics) in Sec. \ref{sec:ass}. After that, we make an in-depth discussion on the comparisons between single-task learning and multi-task learning in APBD tasks, the characteristics of different APBD methods,  the criteria for choosing appropriate APBD methods, etc., followed by extensive APBD related applications, such as crop type classification, yield estimation, stress detection, etc. in Sec. \ref{sec:application}. We envision our promising prospects on APBD domain in Sec. \ref{sec:pros}. Finally, we conclude this review in Sec. \ref{sec:concl}.

\section{Meta-analysis of related literature}
\label{sec:meta}

% As shown in Fig. \ref{fig: overall trend}, the number of APBD-related articles using remote sensing data has increased exponentially since 2019, which is notoriously difficult for those involved in the APBD domain to keep track of this research. To this end, it is essential to periodically conduct a review to summarize recently implemented APBD methods, study areas, tree species, and the types of remote sensing data. In this section, we conduct a meta-analysis regarding the APBD domain to investigate these subjects. 

\begin{figure}[t]
    \centering
    \includegraphics[width=1.0\linewidth]{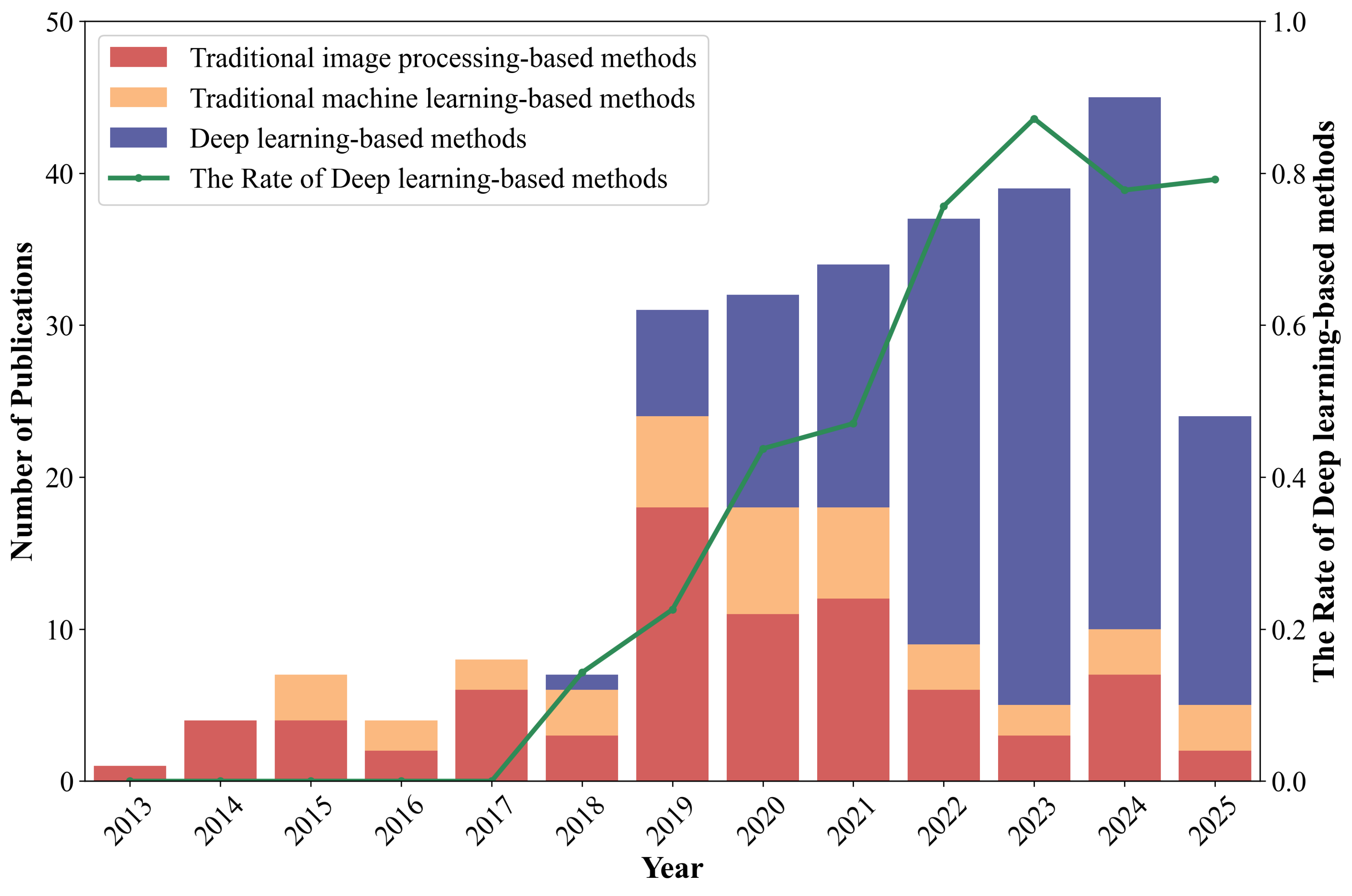}
    \caption{Publication trends of different APBD methodologies from 2013 to 2025.}
    \label{fig: overall trend}
\end{figure}

As illustrated in Fig. \ref{fig: overall trend}, the number of APBD-related studies leveraging remote sensing data has grown exponentially since 2019, making it increasingly challenging for researchers and practitioners in the field to stay abreast of the latest developments. This rapid expansion underscores the need for periodic reviews that synthesize recent advances, including newly implemented APBD methodologies, study areas, targeted crop and tree species, and the types of remote sensing data employed. In this section, we present a meta-analysis of the APBD domain to systematically examine these aspects.

\begin{figure*}[t]
    \centering
    \includegraphics[width=1.0\linewidth]{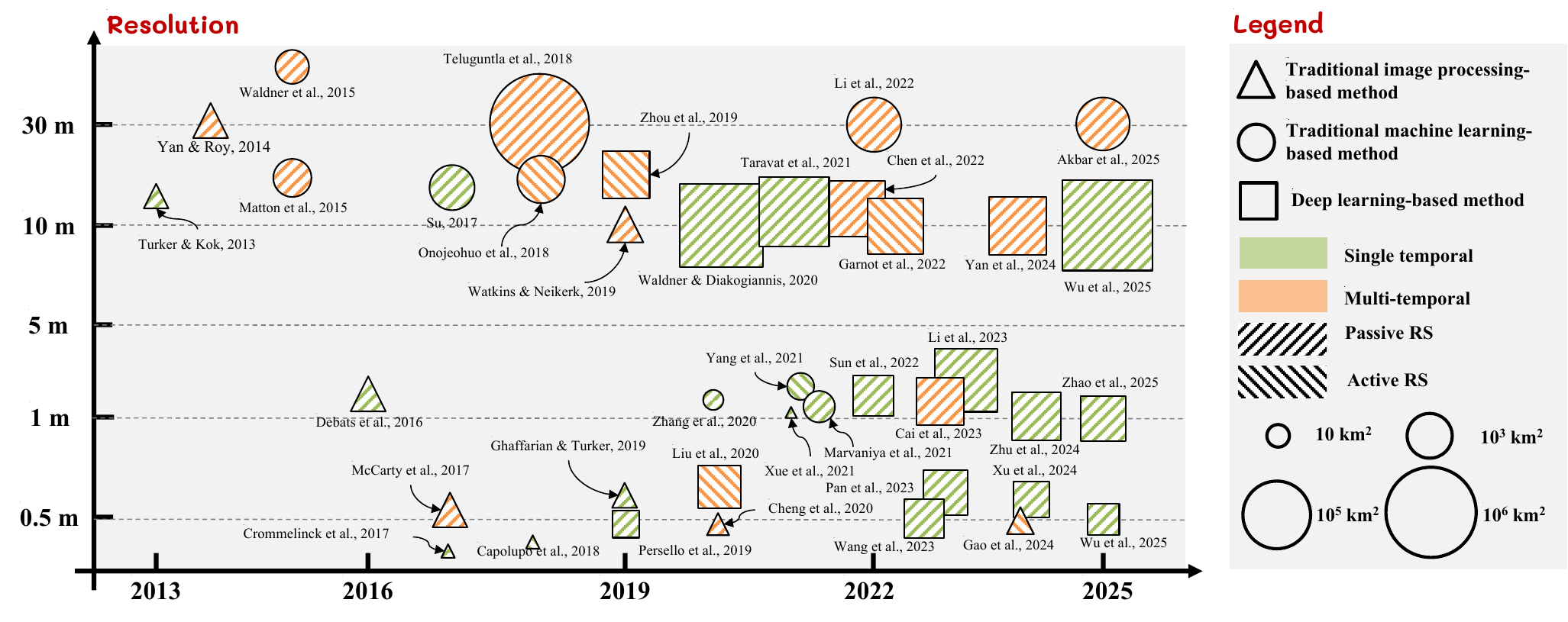}
    \caption{The overall trend of APBD development from some typical examples since 2013. Different shapes represent different APBD methods, and different colors represent the different spatial resolution of optical images. The larger the size, the larger the study area.}
    \label{fig:overallexamples}
\end{figure*}

\subsection{Overall trend of APBD development}

Fig. \ref{fig:overallexamples} presents representative APBD-related studies published between 2013 and 2025. Triangles, circles and rectangles denote traditional image processing-based, traditional machine learning-based, and deep learning-based APBD methods. We can observe some tendencies in the APBD field: 
%Fig. \ref{fig:overallexamples}  Triangles, circles, and rectangles represent methods based on traditional image processing, traditional machine learning, and deep learning, respectively. Several trends can be observed:

\begin{enumerate}
    %\item We can find that the left of Fig. \ref{fig:overallexamples} is sparse, while the right of that holds dense distribution, especially on the bottom-right corner. Since 2019, the number of APBD related researches have been exponentially increased using Very High Resolution (VHR) remote sensing images.
    \item The left side of Fig. \ref{fig:overallexamples} is relatively sparse, whereas the right side, particularly the bottom-right corner, is densely populated. Since 2019, the number of APBD studies using Very High-Resolution (VHR) remote sensing imagery has increased exponentially.
    % \item In terms of the methodology and the limited spatial resolution of data, APBD studies proposed earlier usually use moderate resolution images (such as Sentinel, Landsat, etc.). Some of them adopt VHR images but their study area is really small ($< 10 km^2$). In addition, traditional image processing-based APBD methods are in the majority before 2017. After that traditional machine learning-based APBD methods gradually develop while recently, deep learning-based ITCD methods continuously emerge along with larger-scale study area. 
    \item Early APBD research, constrained by both methodological limitations and the spatial resolution of available data, often relied on moderate-resolution imagery (e.g., Sentinel, Landsat). While some studies employed VHR imagery, their spatial extent was typically small ($< 10\ \mathrm{km}^2$). Before 2017, most methods were based on traditional image processing-based APBD algorithms; traditional machine learning-based APBD methods emerged gradually thereafter, and in recent years, deep learning-based APBD approaches have become dominant, enabling analysis over increasingly large areas.
    \item Prior to 2018, APBD publications generally focused on small study areas (mostly $< 1000\ \mathrm{km}^2$). The growing availability of VHR imagery, advances in computing resources, and the development of more robust AI algorithms have facilitated a shift toward large-scale APBD studies, often integrating multiple VHR datasets.
    \item Although VHR imagery ($< 0.5\ \mathrm{m}$; lower portion of Fig. \ref{fig:overallexamples}) enables high-accuracy APBD, it remains constrained to relatively small areas due to high storage and processing costs. Large-scale studies typically employ imagery with spatial resolution above $10\ \mathrm{m}$, often incorporating multi-temporal datasets to balance accuracy, coverage, and storage demands.
    % \item APBD publications focus on small study area before 2018, most of which are smaller than $10^3 km^2$. In pace with the easier production of VHR remote sensing imagery, the higher performance of computing resources and the more robust artificial intelligence algorithms,  more and more large scale APBD researches have been proposed in recent years using multiple VHR images.
    % \item Although adopting VHR remote sensing data ($<0.5 m$, bottom part in Fig. \ref{fig:overallexamples}) promotes high-accuracy APBD performance, their study area are quite small because of high storage costs. Most of large-scale researches utilize remote sensing data with the spatial resolution over $10 m$, both considering multi-temporal images and lower data storage.
\end{enumerate}

\begin{figure*}[t]
    \centering
    \includegraphics[width=0.95\linewidth]{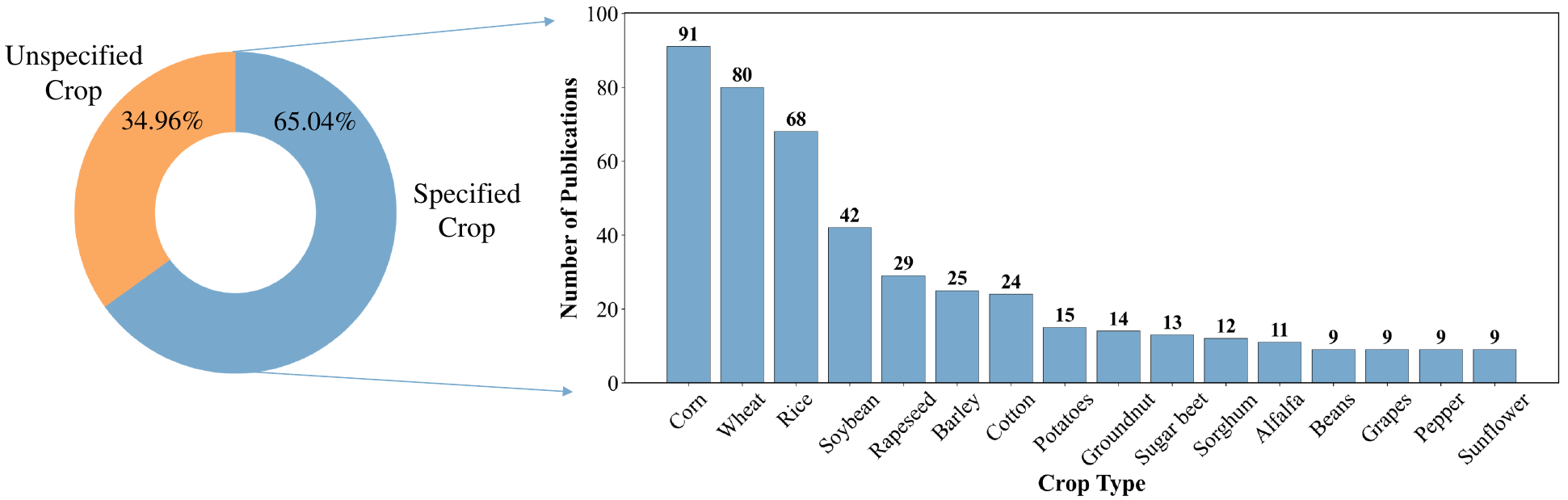}
    \caption{The statistics of crop types in APBD-related publications. In specific crop types, we display only the top 16 most commonly studied types.}
    \label{fig: crop types}
\end{figure*}

\subsection{Quantitative analysis}

% Along with the collection of APBD-related publications, quantitative data are presented through figures in following subsections, including types of cropland, study sites and area, the types of remote sensing data, etc.
Alongside the collection of APBD-related publications, the following subsections present quantitative analyses, covering cropland types, study sites and their spatial extent, sensor types, and spatial resolution.

\subsubsection{Types of cropland}
% Fig. \ref{fig: crop types} presents a statistical summary of the crop types investigated across the reviewed literature. The publications were first categorized based on whether specific crop types were identified, and as illustrated by the pie chart, a substantial majority of publications (65.04\%) focused on specified crops, while a significant portion (34.96\%) did not differentiate between crop types. Due to the wide variety of crops identified in the literature, the accompanying bar chart details the publication frequency for only the top 16 most commonly studied types. The analysis reveals a pronounced research focus on major staple crops, with Corn emerging as the most frequently investigated crop, featured in 91 publications , followed closely by Wheat and Rice with 80 and 68 publications, respectively. Other prominent crops include Soybean (42) , Rapeseed (29) , Barley (25) , and Cotton (24). The list of the top 16 crops also features Potatoes (15) , Groundnut (14) , Sugar beet (13) , Sorghum (12) , Alfalfa (11) , and concludes with Beans, Grapes, Pepper, and Sunflower, each appearing in 9 publications.

Fig. \ref{fig: crop types} provides a statistical overview of the crop types investigated in the reviewed literature. Publications were first classified based on whether they specified particular crop types. As shown in the pie chart, the majority (65.04\%) focused on identified crops, while a considerable proportion (34.96\%) did not distinguish between crop types. Given the wide variety of crops examined, the accompanying bar chart highlights only the 16 most frequently studied. The results reveal a strong emphasis on major staple crops, with Corn being the most extensively investigated (91 publications), followed by Wheat (80) and Rice (68). Other commonly studied crops include Soybean (42), Rapeseed (29), Barley (25), and Cotton (24). The remaining crops in the top 16 include Potatoes (15), Groundnut (14), Sugar beet (13), Sorghum (12), Alfalfa (11), and finally Beans, Grapes, Pepper, and Sunflower, each appearing in 9 publications.

\begin{figure}[t]
    \centering
    \includegraphics[width=1.0\linewidth]{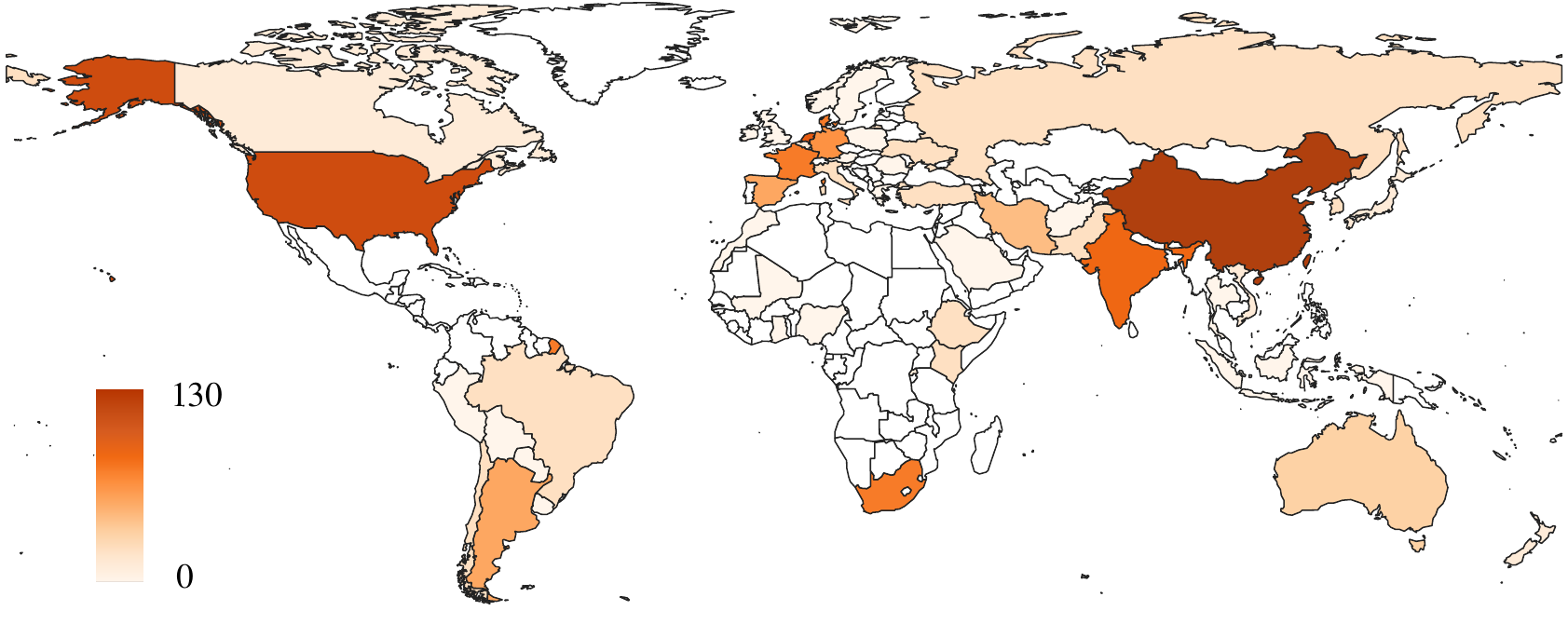}
    \caption{Spatial distribution of global APBD study sites.}
    \label{fig: spatial distribution}
\end{figure}

\subsubsection{Study sites}
%Fig. \ref{fig: spatial distribution} displays the spatial distribution of study sites for Agricultural Parcel and Boundary Delineation (APBD) based on our collected database. The research is geographically concentrated, with a few countries serving as primary study areas. According to the statistics, China is the most prominent research hub with 130 related studies, followed by the United States (19), Netherlands (14), and India (11). Similar to the distribution of research institutions, a majority of these study sites are located in East Asia, North America, and parts of Europe. In contrast, although regions like South America and Africa possess vast agricultural lands and significant research value, the number of studies conducted there is relatively low. This disparity may be attributed to factors such as the complexity of smallholder farming systems, frequent cloud cover affecting optical imagery, and challenges in data acquisition and fieldwork.

Fig. \ref{fig: spatial distribution} illustrates the spatial distribution of study sites for APBD based on our compiled database. The research is geographically concentrated, with a limited number of countries serving as major study areas. China emerges as the leading hub with 130 studies, followed by the United States (19), the Netherlands (14), and India (11). Consistent with the distribution of research institutions, most study sites are clustered in East Asia, North America, and parts of Europe. By contrast, regions such as South America and Africa, despite their extensive agricultural lands and substantial research potential, remain underrepresented. This imbalance may be explained by challenges such as the complexity of smallholder farming systems, persistent cloud cover affecting optical imagery, and difficulties in data acquisition and field surveys.

\subsubsection{Study area}
% Among the literature where the study area size was explicitly reported, an analysis of the spatial scale of research reveals distinct trends over time, as shown in Fig. \ref{fig: study area}. A key observation is that Deep Learning methods are predominantly responsible for the recent increase in studies conducted at very large scales. Notably, the vast majority of publications addressing extremely large regions (Study Area more than 1,000,000 ha) utilize Deep Learning, a trend that becomes particularly pronounced from 2020 onwards. In contrast, while Traditional Image Processing-Based and Traditional Machine Learning methods have been applied across a range of spatial scales, including some large-area applications in earlier years, their use in studies covering the largest geographical extents is significantly less frequent compared to the more recent deep learning-based approaches.

\begin{figure}[t]
    \centering
    \includegraphics[width=1.0\linewidth]{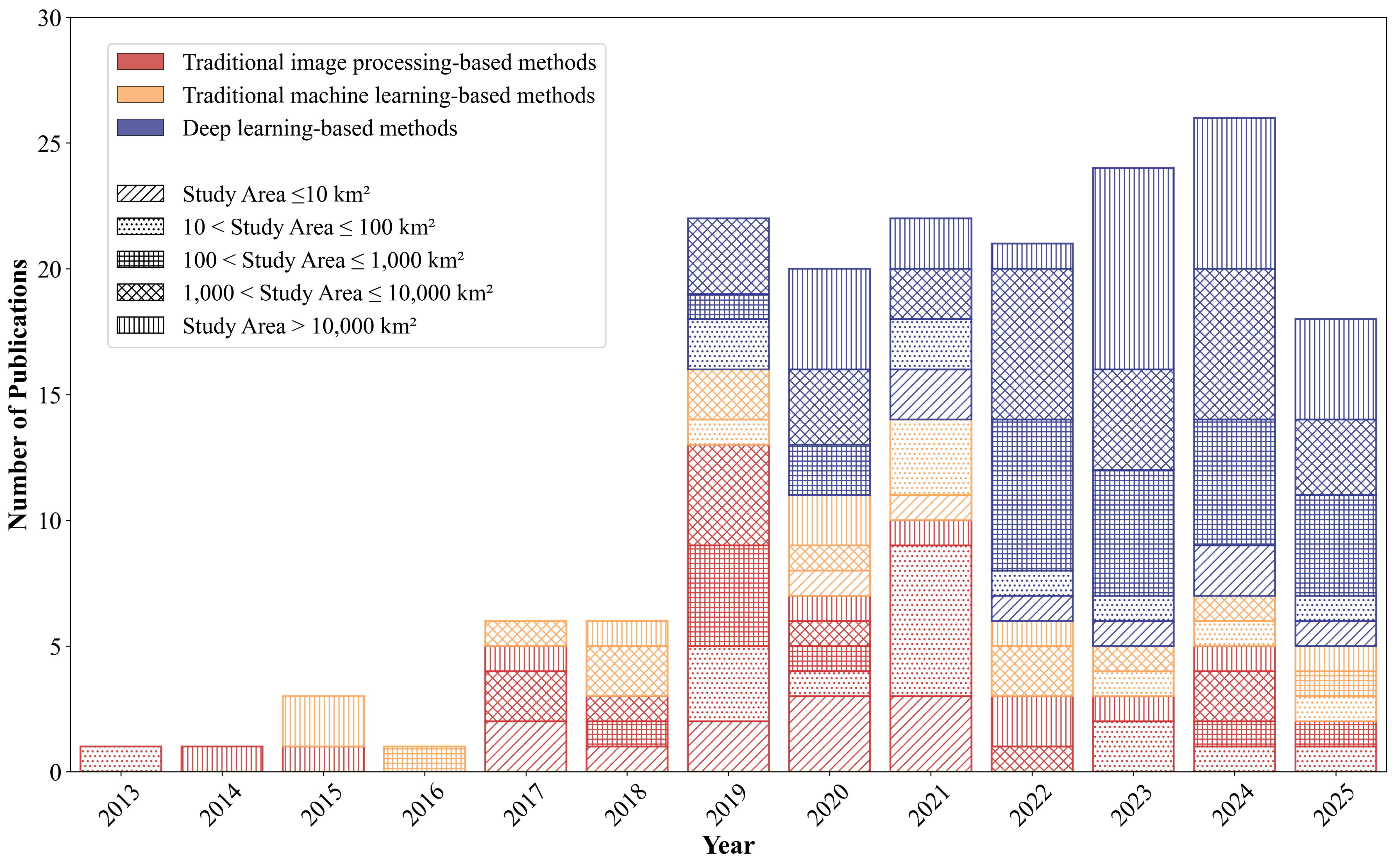}
    \caption{The statistics of study areas in APBD publications. Different textures denote different study areas, and different colors denote different APBD methods.}
    \label{fig: study area}
\end{figure}

Among the studies that explicitly reported their study area size, the analysis of spatial scale over time reveals clear trends (Fig. \ref{fig: study area}). A key finding is that the recent surge in very large-scale research is predominantly driven by Deep Learning methods. In particular, the vast majority of studies covering extremely large regions (over $10,000\ \mathrm{km}^2$) employ deep learning, with this pattern becoming especially evident after 2020. By contrast, while traditional image processing-based and traditional machine learning-based methods have been applied across various spatial scales including some large-area studies in earlier years, their presence in research addressing the largest geographical extents is far less frequent compared to the more recent deep learning-based approaches.

\subsubsection{Sensor type}

% The distribution of sensor types employed in the reviewed literature reveals a strong reliance on satellite-based platforms, as depicted in Fig. \ref{fig: sensor types}. Satellite imagery constitutes the vast majority of data sources, accounting for 84.0\% of the surveyed studies. Data from Unmanned Aerial Vehicle (UAV) and traditional aerial platforms represent smaller but significant proportions, at 10.8\% and 5.2\%, respectively. A detailed analysis of the satellite data shows a relatively balanced utilization of medium-resolution (10m–100m) and high-resolution (<10m) sensors. Among the medium-resolution platforms, Sentinel(82), Landsat (31) are the most prominent. In the high-resolution category, a diverse range of sensors is used, with the GF series (62), WorldView (13), SPOT (7), and Jilin-1 (7) being the most frequently documented.

\begin{figure*}[t]
    \centering
    \includegraphics[width=0.75\linewidth]{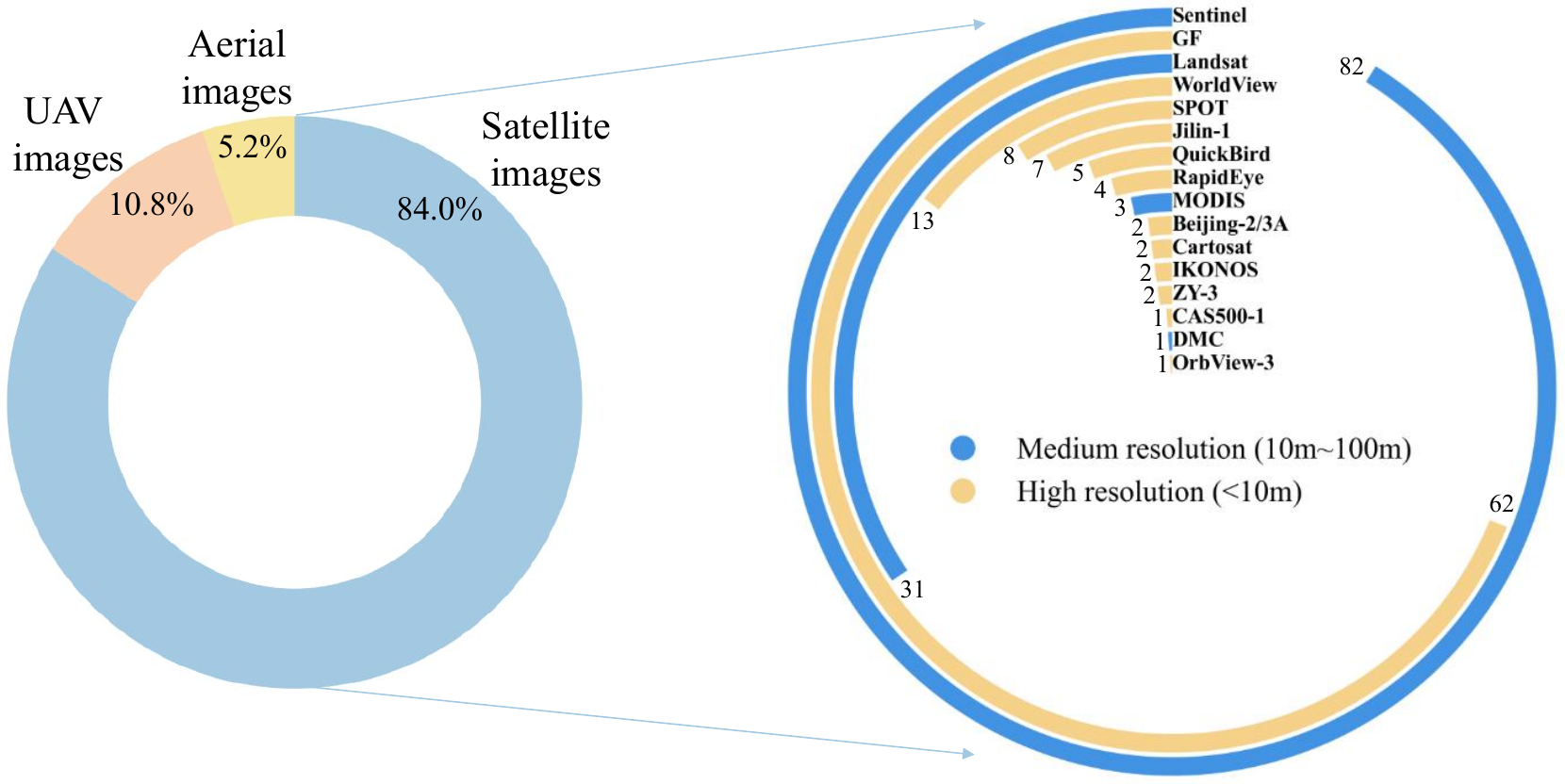}
    \caption{The number of sensor types used in APBD-related publications.}
    \label{fig: sensor types}
\end{figure*}

The distribution of sensor types in the reviewed literature highlights a strong reliance on satellite-based platforms (Fig. \ref{fig: sensor types}). Satellite imagery accounts for the overwhelming majority of data sources, representing 84.0\% of the surveyed studies. Unmanned Aerial Vehicle (UAV) data and aerial imagery contribute smaller yet notable shares, at 10.8\% and 5.2\%, respectively. A closer examination of satellite data indicates a relatively balanced use of medium-resolution (10–100 m) and high-resolution (<10 m) sensors. Within the medium-resolution category, Sentinel (82) and Landsat (31) are the most widely used platforms. In the high-resolution group, a broader range of sensors is reported, with the GF series (62) being the most prominent, followed by WorldView (13), SPOT (7), and Jilin-1 (7).

\subsubsection{Spatial resolution of data}
% An analysis of publications that report the specific spatial resolution of their input data illustrates a clear evolution in data preference over the past decade, as shown in Fig. \ref{fig: spatial resolution}. A predominant trend is the marked increase in the utilization of high-resolution imagery (less than or equal to 1m). This shift has become particularly pronounced in recent years, with a vast number of studies, especially those employing Deep Learning methods, leveraging the fine spatial detail offered by such data. Conversely, while coarser data (more than 10m) has been used across different methodologies, its application within Deep Learning research has shown a noticeable decline in the most recent years. This pattern suggests that as the field advances, there is a growing convergence towards high-resolution data, likely driven by the capacity of modern algorithms like deep neural networks to effectively exploit detailed textural and spatial information for more accurate plot extraction.

\begin{figure}
    \centering
    \includegraphics[width=1.0\linewidth]{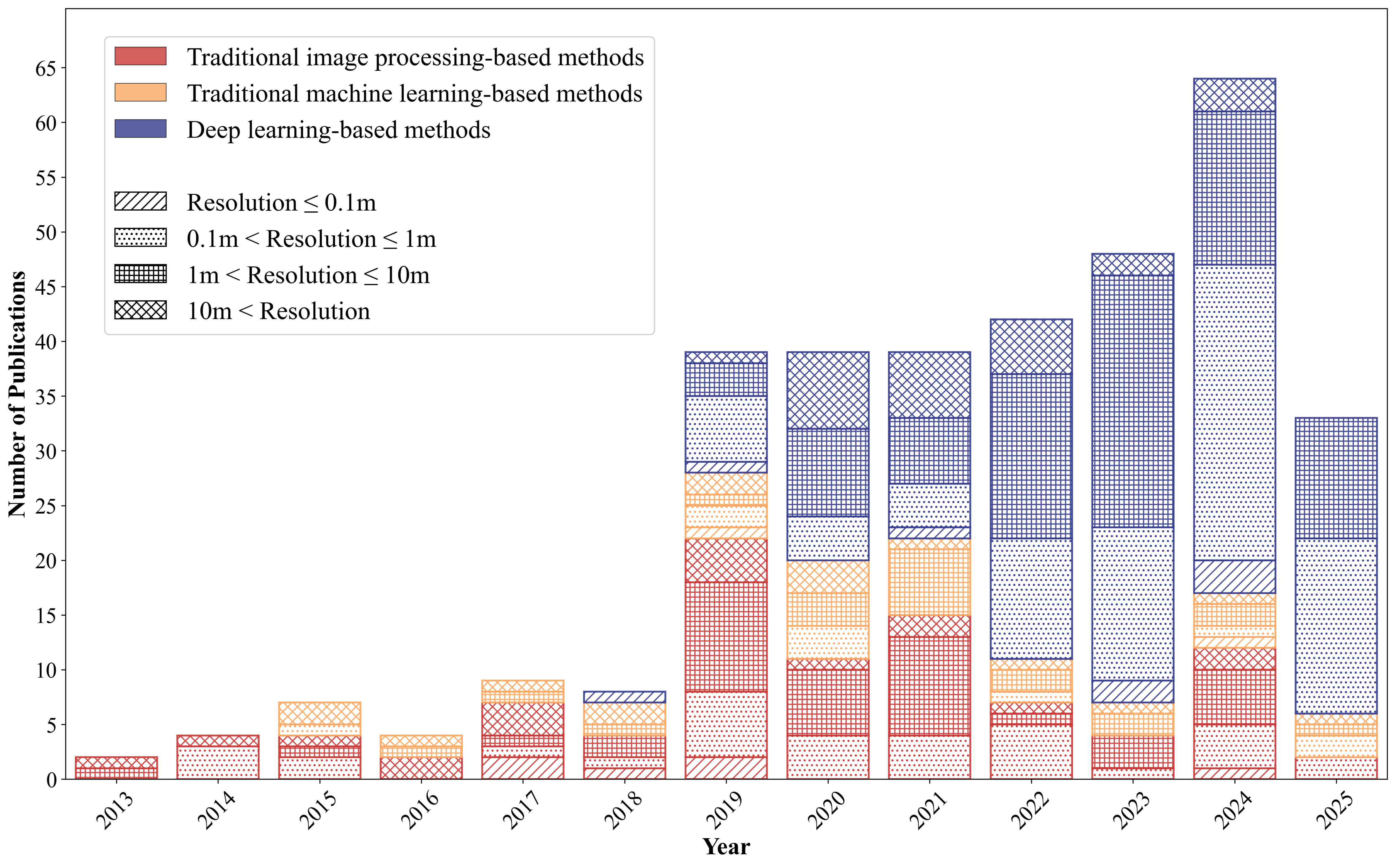}
    \caption{The statistics of spatial resolution of images used in APBD publications. Different textures denote different study areas, and different colors denote different APBD methods.}
    \label{fig: spatial resolution}
\end{figure}

An analysis of publications reporting the spatial resolution of their input data reveals a clear evolution in data preferences over the past decade, as illustrated in Fig. \ref{fig: spatial resolution}. A notable trend is the increasing reliance on high-resolution imagery ($\leq$ 1 m), a shift that has become especially pronounced in recent years. Numerous studies now capitalize on the fine spatial details such data provide, particularly those employing deep learning methods. In contrast, although coarser-resolution data (>10 m) have historically been applied across various methodologies, their use in deep learning research has declined markedly in recent years. This trajectory suggests a growing convergence toward high-resolution datasets, likely driven by the ability of modern algorithms, such as deep neural networks, to effectively harness detailed textural and spatial information for more accurate plot extraction.

\section{Methodology review}
\label{sec:method}

This section reviews the development and summary of APBD methodology. We categorize existing APBD methods into three classes: traditional image processing-based APBD, traditional machine learning-based APBD, and deep learning-based APBD methods. We further categorize existing deep learning-based APBD methods into three subclasses: object detection-based, semantic segmentation-based and foundation model-based APBD methods.

\subsection{Traditional image processing-based APBD methods}

Traditional image processing-based APBD methods include mainly image binarization, detection operators, wavelet transform, watershed segmentation, mean shift and so forth. Inspired by previous segmentation surveys \cite{lucchese2001colour,hossain2019segmentation}, we can categorize them into three major types: pixel-based, edge-based and region-based methods.  
%The former four methods major in tree crown detection tasks, while image segmentation majors in tree crown delineation tasks. 
Table \ref{tab:tip} lists traditional image processing-based APBD methods and the collected examples. Figure  displays some typical examples of traditional image processing-based APBD methods.

\begin{table}[t]
    \centering
    \caption{{The summary of traditional image processing based ITCD methods}}
    %\resizebox{\textwidth}{!}{
    \begin{tabular}{ccc}
    \hline
     Tasks &  {Methods}  &  Examples \\ \hline
     \multirow{2}*{Pixel-based} & {Image binaryzation}   & \cite{da2007delineation,zhu2025integrated} \\ 
     %& Multiple thresholds &  \\
     & Clustering &  \cite{garcia2017machine,ghaffarian2019improved}  \\ \hline
    \multirow{5}*{Edge-based} &  Detection operators  & \cite{watkins2019comparison,robb2020semi}  \\
    &  Wavelet transform & \cite{ji1996delineating,ishida2004application} \\
    & Multiscale combinatorial grouping & \cite{cheng2013high,li2020dbc} \\ 
    & Mean shift & \cite{su2015image,wassie2018procedure} \\ 
    & Globalized probability of boundary & \cite{crommelinck2017contour,marshall2019crowd} \\ \hline
      %\hline
    \multirow{4}*{Region-based} & OBIA \& GEOBIA & \cite{capolupo2018novel,hedayati2022paddy} \\
    & Watershed segmentation & \cite{yan2016conterminous,xue2021watershed} \\
    & Region growing & \cite{mueller2004edge,wagner2020extracting} \\ 
    & Local binary fitting & \cite{li2008minimization,maghsoodi2019development} \\ \hline
    \end{tabular}
    %}
    \label{tab:tip}
\end{table}

\begin{figure*}[t]
    \centering
    \includegraphics[width=1.0\linewidth]{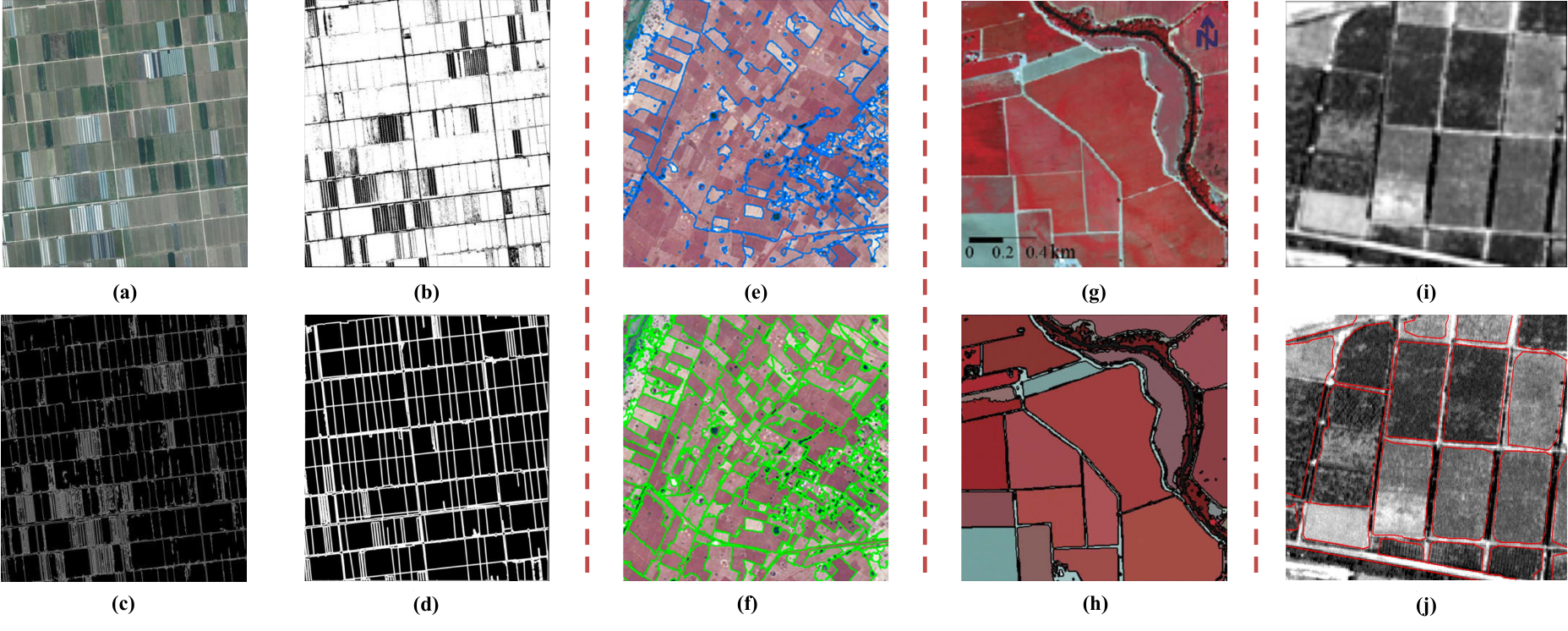}
    \caption{Some typical examples of traditional image processing-based APBD methods. (a) The original aerial image from Hong \textit{et al.} \cite{hong2021development}. (b)-(d) The results of APBD using image binarization, edge detection operator (Canny operator) and Suzuki85 algorithm \cite{hong2021development}. (e)-(f) The results of APBD using globalized Probability of boundary (gPb) and Multi-Resolution Segmentation (MRS) from \cite{marshall2019crowd}. (g) The original image by OrbView-3, located in western California, USA from Su \textit{et al.} \cite{su2015image}. (h) The results of APBD using mean shift algorithm \cite{su2015image}. (i) The original image by Cartosat-1 satellite, located in Tehran, Iran from Maghsoodi \textit{et al.} \cite{maghsoodi2019development}. (j) The results of APBD using the local binary fitting algorithm \cite{maghsoodi2019development}.}
    \label{fig:tip examples}
\end{figure*}

\subsubsection{Pixel-based methods}

Pixel-based methods consist of image binaryzation and clustering segmentation in the feature space \cite{hossain2019segmentation}. In this case, each spatially continuous unit needs to be assigned a unique label. \textbf{Image binaryzation} is a straightforward method to classify pixels or objects that belong to the parcel or the edge by comparing the thresholds. For example, 
%Da Costa \textit{et al.} \cite{da2007delineation} propose a simple thresholding operation according to textural attributes to discriminate between vine field and non-vine field pixels using high-resolution remote sensed images. 
Graesser \& Ramankutty \cite{graesser2017detection} adopt adaptive threshold using temporal Landsat imagery across a South American region. 
Zhu et al. \cite{zhu2025integrated} propose a simple but effective method that convert detected continuous-valued texture results into binary boundaries using thresholding methods. Since the limitations of global thresholding make it challenging to address the issue of coexistence of high-value noise and low-value boundaries in the detection results, they alternatively adopt local thresholding that calculate independent thresholds for each pixel based on neighboring pixels to effectively avoid this issue.
As \textbf{clustering segmentation} methods depend on the features for each pixel, we assign clutering methods to pixel-based APBD methods. There are many classical unsupervised clustering algorithms applied in APBD scenarios, such as $k$-means \cite{garcia2017machine}, fuzzy c-means \cite{ghaffarian2019improved}, ISODATA \cite{rydberg2002integrated}, AUTOCLUST clustering \cite{ma2014cultivated}, Density-Based Spatial Clustering of Applications with Noise (DBSCAN) clustering \cite{li2022machine}, etc.

\subsubsection{Edge-based methods}

Edge-based methods first identify edges and then close them by using contouring algorithms, which assumes that between edges, the pixel properties change abruptly. The main challenge of APBD task is how to accurately detect the edges and boundaries between different parcels. Popular edge-based applied in APBD includes some detection operators, wavelet transform, multiscale combinatorial grouping, mean shift and globalized probability of boundary, etc. 
Traditional \textbf{edge detection operator} mainly involves locating the edge of the local area by using a differential operator to identify the place where the gray value abruptly changes in the image, such as Canny operator \cite{turker2013field}, Sobel operator \cite{duvvuri2023hs}, Roberts operator and Scharr operator \cite{watkins2019comparison}, etc., which are also called gradient operators. However, the extraction effect of the simple traditional edge detection operators is usually not ideal \cite{wang2023survey}. 
\textbf{Wavelet transform} evolves from Fourier analysis, which has the advantages of multi-resolution analysis and has good approximation characteristics for one-dimensional signals. For instance, Ishida \textit{et al.} \cite{ishida2004application} use multi-resolution wavelet transform from SPOT image to detect the edges of submerged paddy fields, which performs better than the Difference of Gaussian (DoG) filter. 
\textbf{Globalized Probability of boundary} (\textbf{gPb}) \cite{arbelaez2010contour} combines the steps of image segmentation, line extraction, and contour generation with the ability to integrate image information at both local and global scales for texture, color, and brightness, which is an extremely advanced edge detection algorithm. For example, Marshall \textit{et al.} \cite{marshall2019crowd} utilize gPb to achieve to determine farm boundaries from WorldView imagery in highly fragmented agricultural landscapes of Ethiopia. In addition, gPb is also applied in building boundary extraction \cite{chen2020automatic}. 
\textbf{Multiscale combinatorial grouping} generates each boundary and region based on structural forest edge detector, spectrum division and global weighting, which usually has higher boundary extraction accuracy than the gPb and the classical edge detection Sobel algorithm \cite{arbelaez2010contour}. 
\textbf{Mean shift} is a nonparametric density estimation algorithm that eliminates the need to assume a sample distribution model and determine the number of categories, which can effectively reduce intra-field variation and preserve edge information. Su et al. \cite{su2015image} propose a new image segmentation algorithm based on mean shift to single out croplands in high-resolution remote sensing imagery.

\subsubsection{Region-based methods}
\label{sec:tip_region}
The region-based methods start from the inside of an object and then expand outward until meeting the object boundaries, assuming that neighboring pixels within the same region have similar values. Commonly, region-based methods contains Object-Based Image Analysis (OBIA) or GEographic Object-Based Image Analysis (GEOBIA), watershed segmentation, region growing and local binary fitting, etc. 
\textbf{OBIA} or \textbf{GEOBIA} \cite{zhang2013boundary} is a sub-discipline of GIScience devoted to partitioning remote sensing imagery into meaningful image-objects and assessing their characteristics through spatial, spectral and temporal scale, which could analyze high-spatial resolution imagery by using spectral, spatial, textural and topological characteristics. Capolupo \textit{et al.} \cite{capolupo2018novel} propose a method for automatic extraction of terraces from historical and contemporaneous aerial photos based an OBIA approach. 
\textbf{Watershed segmentation} is a mathematical morphological segmentation algorithm that is based on topology theory, simulating the process of water immersion \cite{vincent1991watersheds}. For example, Xue \textit{et al.} \cite{xue2021watershed} propose an improved watershed segmentation algorithm based on a combination of pre- and post-improvement procedures using Gaofen-2 remote sensing image with high time efficiency and segmentation accuracy. The watershed algorithm can obtain accurate boundary results for single-pixel localization; however, it is prone to over-segmentation. 
\textbf{Region growing} is the most popular and simple algorithms for region-based segmentation, which includes two main issues: selection of seed region and similarity \cite{adams1994seeded}. Wagner \& Oppelt \cite{wagner2020extracting} propose a graph-based growing contours algorithm that is capable of extracting complex networks of boundaries present in agricultural landscapes, and is largely automatic with little supervision required in the northern Germany using multi-temporal Sentinel-2 imagery. 
\textbf{Local binary fitting} is an energy-based segmentation algorithm that effectively segments images with uneven grayscale with well-performance in satellite imagery of flatter farmland. For instance, Maghsoodi \textit{et al.} \cite{maghsoodi2019development} advance the local binary fitting model as a multi-phase model with nonparametric active contours, followed by adding two texture layers to the input images and the development of the external energy function.

Besides abovementioned region-based algorithms, other segmentation methods, such as {Multi-Resolution Segmentation (MRS)} \cite{song2019object,tetteh2021evaluation,borowiec2022using}, {spatial-temporal information segmentation} \cite{garcia2018outlining,cheng2020destin} and {Variational Region-based Geometric Active Contour} ({VRGAC}) \cite{yan2014automatic,yan2016conterminous}, Suzuki85 Algorithm \cite{suzuki1985topological,hong2021development} etc. are also applied in many APBD scenarios. Many studies improve the original segmentation algorithms based on computer vision domain. For example, the MRS algorithm is a well-known method for segmenting objects from images, but the quality of segmentation depends on the a priori knowledge of which scale, shape and compactness values to use. Tetteh \textit{et al.} \cite{tetteh2020optimal} propose a sequential model-based optimization method based on MSR, which outperforms other segmentation optimization methods.
Furthermore, some APBD-based researches combined different traditional image processing-based APBD methods to simultaneously improve the results. For example, Watkins \& Niekerk \cite{watkins2019comparison} combined canny edge detection operator and watershed image segmentation algorithm to produce the accurate field boundaries with an OA of 92.9\%. Xu \textit{et al.} \cite{xu2019farmland} applied mean shift and multi-resolution segmentation using GF-2 and QuickBird images to benefit the accurate farmland information extraction. Borowiec \& Marmol \cite{borowiec2022using} propose a hybird framework to identify agricultural land boundaries using LiDAR data. They combine operators of Prewitt and Canny and the multi-resolution segmentation, along with PCA and Hough Transform to attain precise determination of agricultural land boundaries.

\subsection{Traditional machine learning-based APBD methods}

The revolution in machine learning facilitates the development of APBD by offering powerful, adaptable, and accurate solutions. Generally speaking, for both agricultural parcel detection and boundary delineation, there are four steps in traditional machine learning-based APBD methods: 1) image preprocessing, 2) feature extraction, 3) classifier training, and 4) model prediction. Here we focus more on the nature of APBD, which is progressing in feature extraction and classifier training. Table \ref{tab:tml} lists traditional machine learning-based APBD methods and the collected examples. This section does not separate agricultural parcel detection and boundary delineation, because feature extraction and classifier training are both necessary for them and the employed methods are similar.

\begin{table}[t]
    \centering
    \caption{{The summary of feature extraction and adopted classifiers in the traditional machine learning-based APBD methods}}
    %\resizebox{\textwidth}{!}{
    \begin{tabular}{ccc}
    \hline
       Items  & Methods & Examples \\ \hline
        & \multirow{2}*{Non-handcrafted features}  & \cite{debats2016generalized,onojeghuo2018mapping} \\
       Feature &    & \cite{zhang2020parcel,zheng2024farmland} \\
       extraction & \multirow{2}*{Handcrafted features}  & \cite{valero2016production,su2017efficient} \\
       & & \cite{li2020svm,cai2022adaptive} \\
       \hline
       
        & Decision tree & \cite{garcia2017machine,tariq2023mapping} \\
        & Classification and Regression Trees (CART) & \cite{watkins2019automating,bhavana2024crop} \\ 
       \multirow{2}*{Adopted} & Support Vector Machine (SVM) & \cite{li2020svm,savitha2023mapping} \\
        \multirow{2}*{classifiers} & Genetic Programming (GP) & \cite{lu2022genetic,wen2022object} \\
         & Random Forests (RFs) &  \cite{teluguntla201830,cai2022adaptive}  \\      
       & Maximum likelihood classifier & \cite{abou2003improvements,matton2015automated} \\   
       & Dynamic Time Warping (DTW) & \cite{belgiu2018sentinel,mondal2018mountain} \\ \hline

    \end{tabular}
    %}
    \label{tab:tml}
\end{table}

\subsubsection{Feature extraction}

There is a variety of feature extraction methods, which can be simply classified into two types, i.e., non-handcrafted features and handcrafted features. Non-handcrafted features utilize mainly the obvious inner features of images themselves, such as spectral information \cite{estes2022high,savitha2023mapping}, vegetation index \cite{onojeghuo2018mapping,mashaba2021delineating}, texture characteristics \cite{garcia2017machine,zheng2024farmland}, geometrical features \cite{debats2016generalized,zhang2020parcel}, climate variables \cite{akbar2025multi}, and so forth. Some studies also take homogeneity factors \cite{yang2025enhanced}, temporal patterns \cite{waldner2015automated,belgiu2018sentinel,luo2021using} and point cloud data into consideration. 
On the other hand, handcrafted features are specific image representations that are crafted by domain knowledge and prior understanding of the data. These features are created by specific methods (e.g., gradient direction histogram \cite{li2020svm}, gray level co-occurrence matrix \cite{cai2022adaptive}, differential of Gaussian \cite{su2017efficient}, principal component analysis \cite{valero2016production} and so on) to capture relevant information that is deemed important for APBD tasks.

The interpretability of these features makes them useful for understanding and reasoning about the content as they are explicitly designed to capture certain visual attributes like shapes and edges of agricultural parcels. Compared to non-handcrafted features, handcrafted features present more data-driven characteristics. Still, due to the requirements of manual design and expert understanding, these features lack scalability and transferring ability to new scenarios. In a nutshell, a full understanding of the characteristics and the specific demands of specific APBD tasks is essential to harness the full potential of these features and is beneficial for later classifier training.

\subsubsection{Classifier training}

Classifier training is the most important part of traditional machine learning-based APBD methods. Potential classifiers contain Decision Trees (DTs), Gaussian maximum likelihood,
linear discriminant analysis, Support Vector Machines (SVMs), Bayesian optimization, Random Forests (RFs), Multi-Layer Perceptrons (MLPs), K-Nearest Neighbors (KNN), Classification and Regression Trees (CART), logistic regression, and so forth. 
Rahman \textit{et al.} \cite{rahman2019season} investigate and test six different classification algorithms to select the best algorithm for Landsat scenes between May and mid-August for USA. The experimental results indicate that RFs achieves the best accuracy, followed by the KNN, SVM, DT, MLP and Gaussian Naive Bayes.
In addition, some researcher adopt multi-model ensemble learning that combines the outputs of different machine learning classifiers. For example, Akbar \textit{et al.} \cite{akbar2025multi} develop an ensemble learning model including RFs, SVM, KNN, gradient tree boost and CART in Google Earth Engine to achieve high-fidelity, high-resolution (30m) annual maps of irrigated areas from 2007 to 2022 in the Upper Red River Basin, U.S., with improved the ground truth accuracy to 84\%. 

In summary, the performance of traditional machine learning-based methods in APBD relies on efficient feature extraction and powerful classifier training. Compared to traditional image processing-based methods, the ability to automatically learn and extract relevant features from raw data that involve specific expert understanding of traditional machine learning-based methods makes them more adaptable to more different scenarios for APBD. However, it is important that a set of high-quality and high-quantity input data are a sufficient condition for the promising performance of traditional machine learning-based methods. 
For example, if the study area is a small region with simple tree targets and landscape invariance and the images are full of noise, traditional image processing-based methods may have better performance. Therefore, choosing or comparing traditional image processing-based and traditional machine learning-based methods, depends on the specific scenarios and data conditions. 
Also, to improve APBD results with limited labels, Estes \textit{et al.} \cite{estes2022high} utilize active learning. They create a platform that rigorously assesses and minimizes label error, and used it to iteratively train a RFs classifier with active learning, which identifies the most informative training sample based on prediction uncertainty.

Furthermore, some studies combine traditional image processing-based and traditional machine learning-based APBD methods to simultaneously improve the performance of APBD scenarios. For example, Yang \textit{et al.} \cite{yang2025enhanced} propose an object-oriented multi-scale segmentation method combined with a SVM, leveraging spectral reflectance, texture, and temporal differences between farmland and non-farmland plots. This approach effectively improves farmland plot extraction accuracy, supporting crop type identification and advancing digital agricultural management. 
Belgiu \& Csillik \cite{belgiu2018sentinel} firstly adopt traditional image processing-based method (MRS) and then adopt two kinds of classifiers (i.e., Time-Weighted Dynamic Time Warping (TWDTW) and RFs) using time-series Sentinel-2 data. Experimental results show that TWDTW achieved comparable classification results to RFs in Romania and Italy, but RFs achieved better results in the USA.

\begin{table}[t]
    \centering
    \caption{{The summary of the deep learning based ITCD methods}}
    %\resizebox{\textwidth}{!}{
    \begin{tabular}{ccc}
    \hline
       Methods  & Networks & Examples \\ \hline
       
         &  U-Net  & \cite{xu2022delineation,wang2023mde} \\
       & DeepLab   & \cite{wang2023bsnet,liu2022deep} \\
        & PSPNet  &  \cite{zhang2020generalized,zhu2024generalized}   \\
       Semantic &FPN &  \cite{xu2023deriving,zhang2024parcel} \\
       segmentation-based & HRNet  & \cite{xie2023edge,yang2025high}  \\
       & D-LinkNet & \cite{luo2023mlgnet,zhang2025toward} \\
       & FCN & \cite{persello2019delineation,sun2022deep} \\
       & ResNet-like  & \cite{huan2021maenet,onojeghuo2023deep} \\
       \hline
       
       \multirow{4}*{Object detection-based} & YOLO   &  \cite{wang2024aams,kim2025development} \\
       & Mask R-CNN   &  \cite{lv2020delineation,tetteh2023comparison} \\
       & Mask2Former &  \cite{zhong2023multi,liu2024edge} \\
        & DeepSnake & \cite{pan2023e2evap,xu2024multiscale} \\
       \hline

       \multirow{5}*{Transformer-based} &  ViT & \cite{xu2024utilizing,zhao2024irregular} \\
       & Swin Transformer & \cite{xu2023evaluation,wu2025parcel} \\
       & SegFormer  & \cite{chen2024novel,wu2025sbdnet}  \\
       %& Relationformer & \cite{xia2024crop} \\
       &  SAM   & \cite{long2024integrating,sun2025sidest} \\
         & VLM & \cite{wu2025fsvlm,wu2025farmseg_vlm} \\
       \hline
       
    \end{tabular}
    %}
    \label{tab:tdl}
\end{table}

\begin{figure}
    \centering
    \includegraphics[width=1\linewidth]{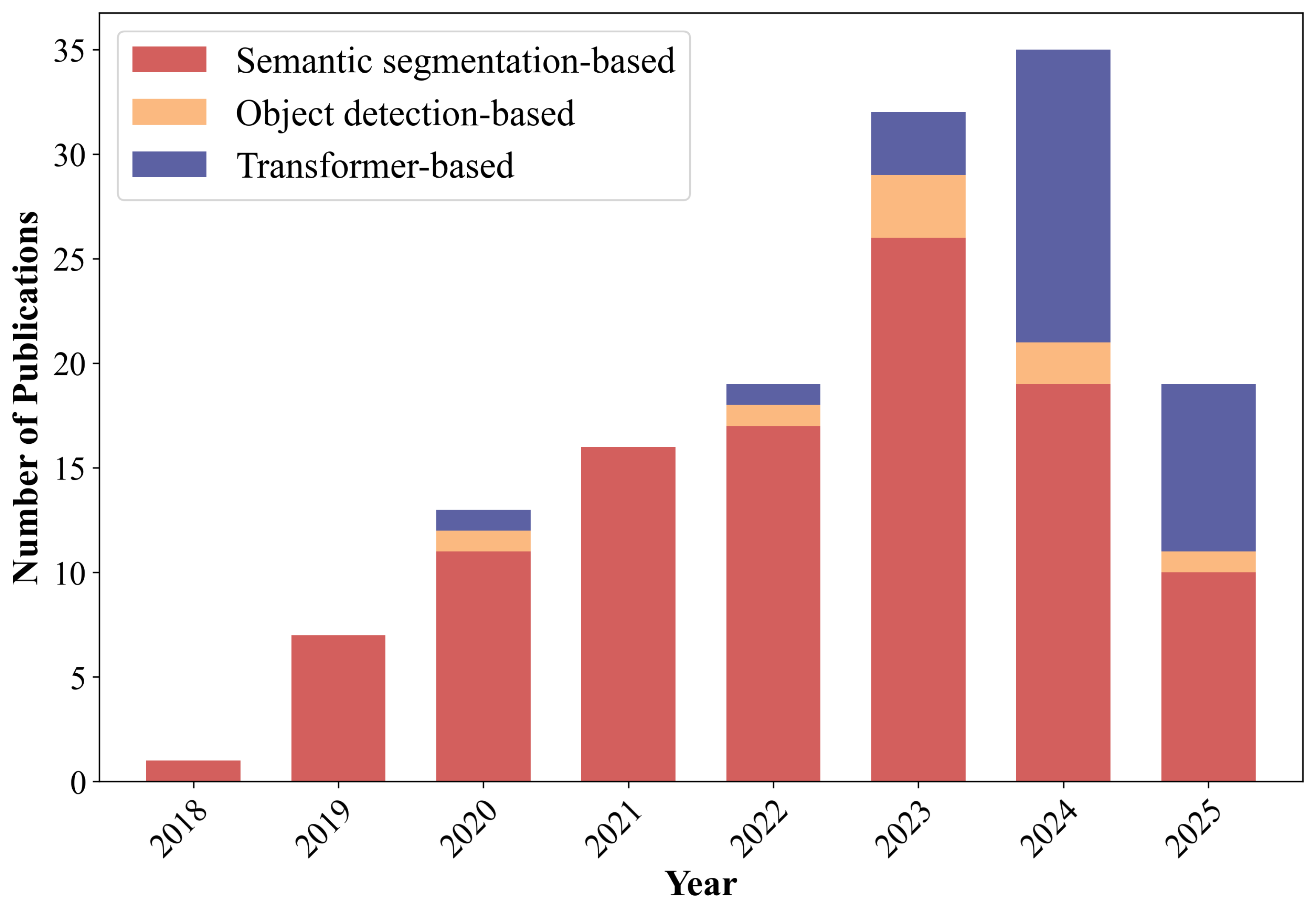}
    \caption{The number of deep learning-based APBD methods-related publications since 2018}
    \label{fig:deep_learning_methods}
\end{figure}

\begin{figure}[t]
    \centering
    \includegraphics[width=1.0\linewidth]{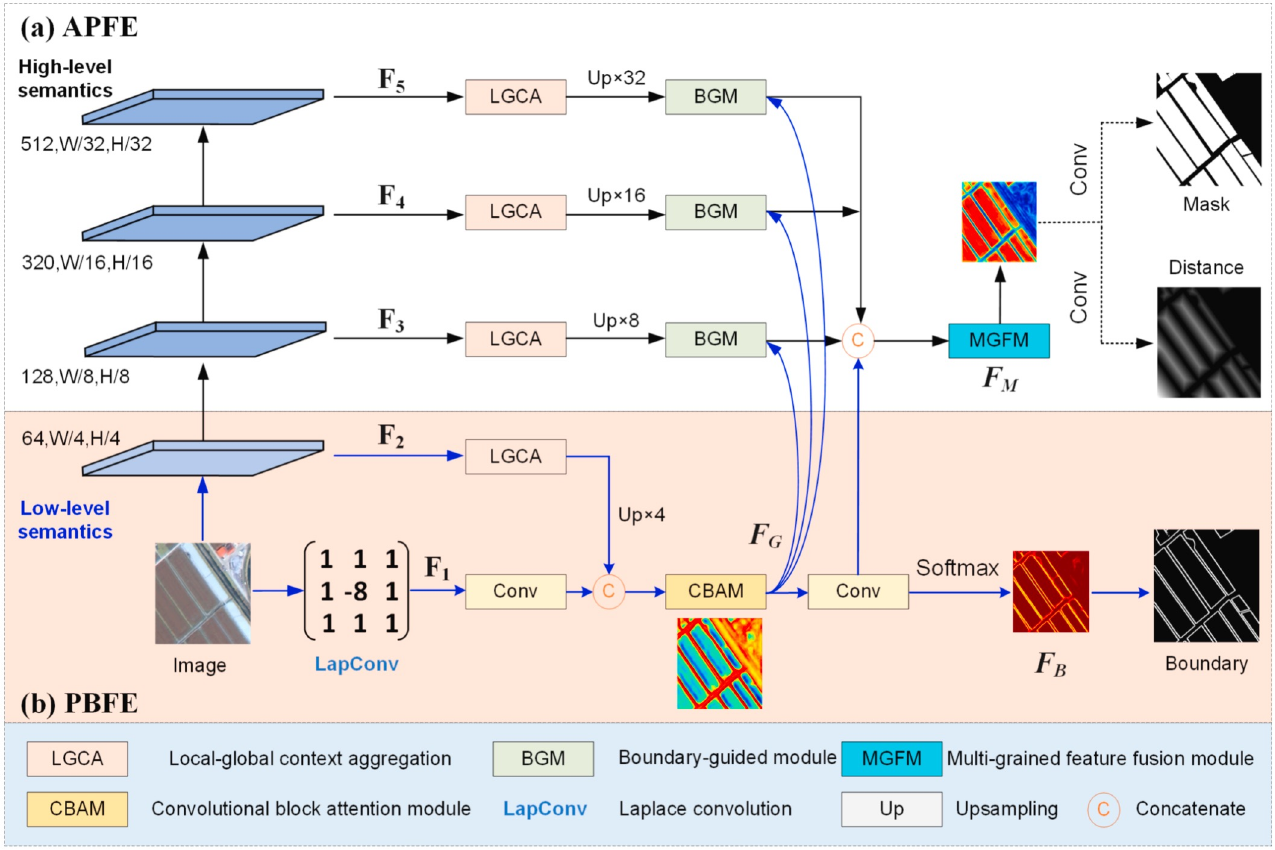}
    \caption{A typical example of a semantic segmentation-based APBD method named HBGNet proposed by Zhao \textit{et al.} \cite{zhao2025large}.}
    \label{fig:dl_ss}
\end{figure}

\subsection{Deep learning-based APBD methods}

As successful cases emerge in various applications, today, many APBD methods adopt neural networks, achieving high-accuracy and real-time APBD results in complex and large-scale regions. Here we review deep learning-based APBD methods by an extended taxonomy: object
detection-based methods, semantic segmentation-based methods and transformer-based methods for agricultural parcel extraction. Table \ref{tab:tdl} lists deep learning-based APBD methods. Figure \ref{fig:deep_learning_methods} displays the number of deep learning-based APBD methods related publications since 2018.

There are many public dataset that could perform crop identification task, which is the simplest task in APBD, such as UC Merced \cite{yang2010bag}, AID \cite{xia2017aid}, NWPU-RESISC45 \cite{cheng2017remote}, etc., and many researches have applied these public dataset or propose new image-level dataset \cite{nascimento2023productive} in many agricultural applications using deep learning algorithms to identify whether an image belongs to crop field. However, they focus on scene classification, which is a coarse crop mapping method using time-consuming sliding window-based scheme and beyond the scope of this review.

\subsubsection{Semantic segmentation-based APBD methods}

Without requiring the time-consuming sliding-window scheme, the semantic segmentation-based APBD method is an end-to-end algorithm, which is the most popular scheme kind of deep learning-based APBD methods. Dissimilar to the object detection-based methods that produce one label for a patch of an image, semantic segmentation methods aim at generating dense classes for each pixel in the whole image. Similar to object detection-based methods, semantic segmentation-based APBD methods also derive from semantic segmentation methods for natural images. Some state-of-the-art semantic segmentation architectures, such as DeepLab \cite{chen2017deeplab}, U-Net \cite{ronneberger2015u}, Fully Connected Networks (FCNs), High-Resolution Network (HRNet) \cite{wang2020deep} and Pyramid Scene Parsing Network (PSPNet) \cite{zhao2017pyramid}, and so forth, have been applied to the APBD domain in recent years. 
Some researchers have employed semantic segmentation-like models to generate confidence maps for agricultural parcel extraction \cite{li2024comprehensive}, where pixels with high confidence indicate the locations of agricultural parcels. 
Many studies have also proposed modified existing semantic segmentation models (like U-Net) for delineation tasks \cite{liu2025lightweight} or simultaneously use multi-task learning methods to improve boundary delineation.
For example, Li \textit{et al.} \cite{li2023using} presents a Semantic Edge-Aware multi-task neural Network (SEANet) to obtain closed boundaries, which integrates three correlated tasks: mask prediction, edge prediction, and distance map estimation. Experiments show that SEANet is an accurate, robust, and transferable method for various areas and different remote sensing images. Xu \textit{et al.} \cite{xu2023extraction} develop a multi-task network model to extract plot-level cropland information, consisting of a cascaded multi-task network with integrated semantic and edge detection, a refinement network with fixed edge local connectivity, and an integrated fusion model. 
Cai et al. \cite{cai2023improving} propose a Dual branch Spatiotemporal Fusion Network (DSTFNet) that integrated very high-resolution images and medium-resolution satellite image time series to extract agricultural field parcels over various landscapes by exploiting important spectral, spatial and temporal information from multi-modal satellite data.
Some researchers have explored the transferability of semantic segmentation-based APBD methods. For instance, Liu \textit{et al.} \cite{liu2022deep} and Tian \textit{et al.} \cite{tian2024fieldseg} propose FieldSeg-DA and FieldSeg-DA2.0, respectively to improve the performance of cross-regional and cross-temporal accurate boundary localization and parcel extraction, which borrow ideas from Fine-grain Adversarial Domain Adaption (FADA).

% For comparison among different semantic segmentation methods, Ochoa and Guo [122] evaluate five state-of-the-art tree delineation methods for segmenting citrus trees from UAV multispectral images: DeepLabV3+, dynamic dilated convolution network (DDCN), SegNet, U-Net, and FCN. The experimental results showed comparable F1 scores, with DDCN achieving the highest F1 score: 94.42\%. However, DDCN exhibited the lowest detection efficiency, taking 1.02 min per hectare, while other algorithms processed each hectare in approximately 15 s.

As the spatial complexity and heterogeneity of features resulting from high resolution makes it difficult to obtain parcel-level information quickly and accurately, some semantic segmentation-based APBD methods may utilize traditional image processing-based APBD method to improve the low-level feature extraction and achieve better boundary delineation performance. For example, Hierarchical semantic Boundary-Guided Network (HBGNet) \cite{zhao2025large} proposes a novel multi-task learning framework, which employs a module based on classical Laplace convolution operator to enhance the model’s awareness of parcel boundary (see details in Fig. \ref{fig:dl_ss}). 
Zhu \textit{et al.} \cite{zhu2024generalized} utilize oriented watershed transformation and hierarchical region merging to address weak boundary loss from modified PSPNet \cite{zhang2020generalized}, leveraging the observational hierarchy of fields, resulting in stable parameters across regions and models. 
Furthermore, some semantic segmentation-based APBD methods firstly employ deep learning-based method to delineate the agricultural boundaries, and then use traditional machine learning method to exactly classify the crop types in each parcel. For example, Tang \textit{et al.} \cite{tang2025parcel} firstly apply SEANet \cite{li2023using} to delineate agricultural land parcel boundaries precisely. Subsequently, a parcel-scale RFs model is developed to enable accurate crop type classification and spatial delineation of crop planting structures.

\subsubsection{Object detection-based APBD methods}

\begin{figure}[t]
    \centering
    \includegraphics[width=1.0\linewidth]{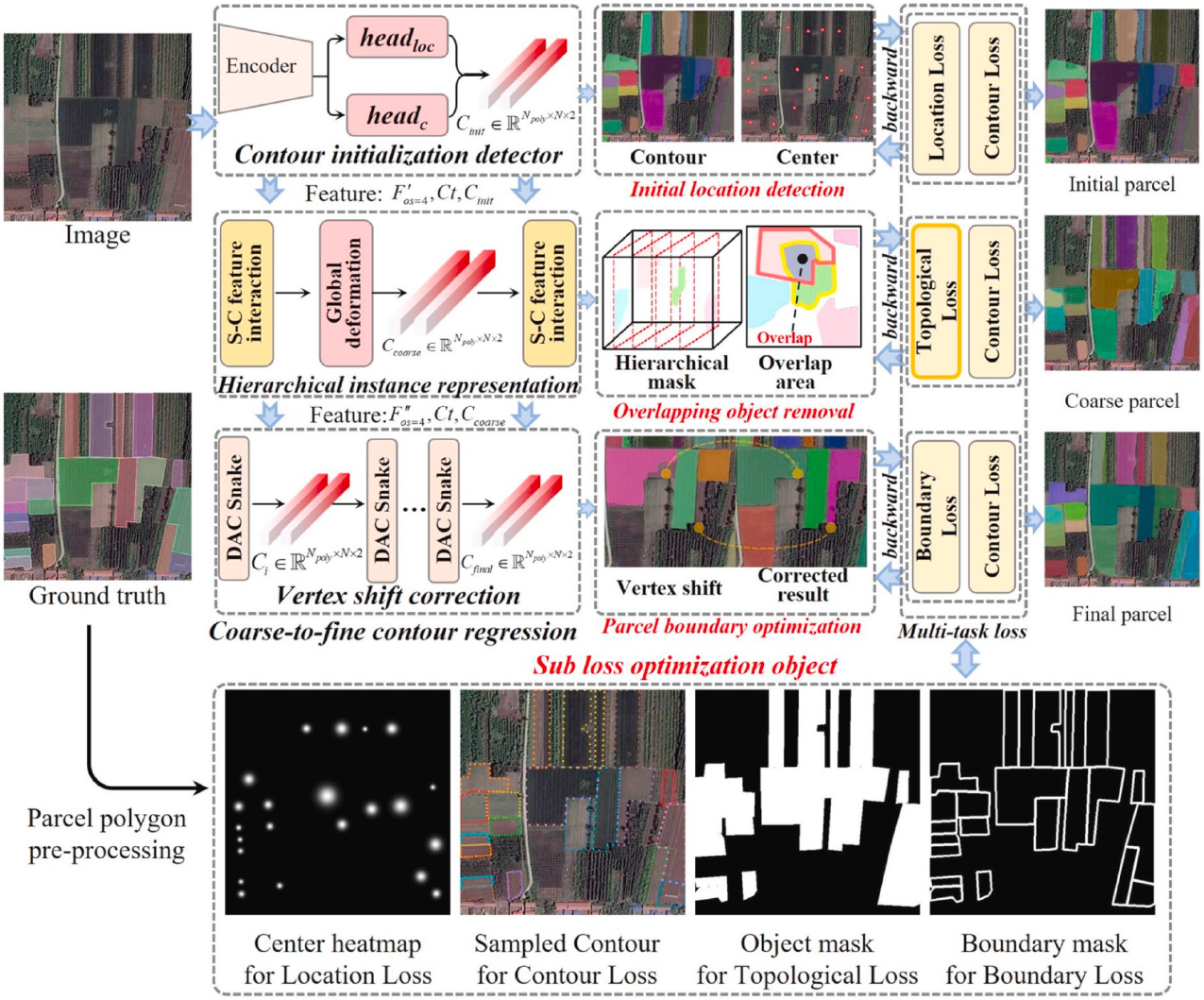}
    \caption{A typical example of an object detection-based APBD method named E2EVAP proposed by Pan \textit{et al.} \cite{pan2023e2evap}.}
    \label{fig:dl_od}
\end{figure}

Object detection-based APBD frameworks, such as YOLO \cite{redmon2016you}, Mask R-CNN \cite{he2017mask}, Mask2Former \cite{cheng2022masked}, DeepSnake \cite{peng2020deep} and so forth, provide precise identification of agricultural parcel and boundary features, especially in farmland with complex and fragmented boundaries. The core of object detection based methods is the region proposal before feature extraction and segmentation. The selective search algorithm extracts approximately thousands of class-agnostic region proposals from an input image, focusing on the complete patches rather than pixels. By combining region generation with refined segmentation, these methods effectively address features such as cropland patches and agricultural boundaries. For example, inspired by DeepSnake, Pan \textit{et al.} \cite{pan2023e2evap} propose the end-to-end vectorization of smallholder agricultural parcel boundaries framework for extracting the vertices of each parcel boundary individually in smallholder farming regions, where the semantic-contour interaction and topological loss through hierarchical instance representation are designed for aggregating the foreground features and jointly establishing the topological relationship between instances to alleviate the topological overlap between parcel objects (See details in Fig. \ref{fig:dl_od}. This method shows considerable gains on the iFLYTEK public agricultural parcel dataset compared to other object detection-based methods (such as Mask R-CNN, E2EC and DeepSnake). 
Cao et al. \cite{cao2023case} propose an example segmentation method of Mask R-CNN based on dual attention mechanism feature pyramid network to describe small farms, where the standard edge detection algorithm cannot accurately segment the blurred farmland boundary. Experiments indicate that this method could accurately depict small farms in very high resolution satellite images, which lays a foundation for the automatic segmentation of small farms.

However, object detection-based method commonly performs a little worse than semantic segmentation-based method. Tetteh \textit{et al.} \cite{tetteh2023comparison} compare one object detection-based method (\textit{i.e.}, Mask R-CNN) with two semantic segmentation-based methods (\textit{i.e.}, U-Net and ResUNet) and one traditional image processing-based method (\textit{i.e.}, MRS based on supervised Bayesian optimization) for delineating agricultural parcels in Lower Saxony, Germany from Monotemporal Sentinel-2 images. Experimental results show that ResUNet combined with a post-processing approach generated the best segmentation results, closely followed by the optimized MRS approach.

\subsubsection{Transformer-based APBD methods}

\begin{figure}[t]
    \centering
    \includegraphics[width=1.0\linewidth]{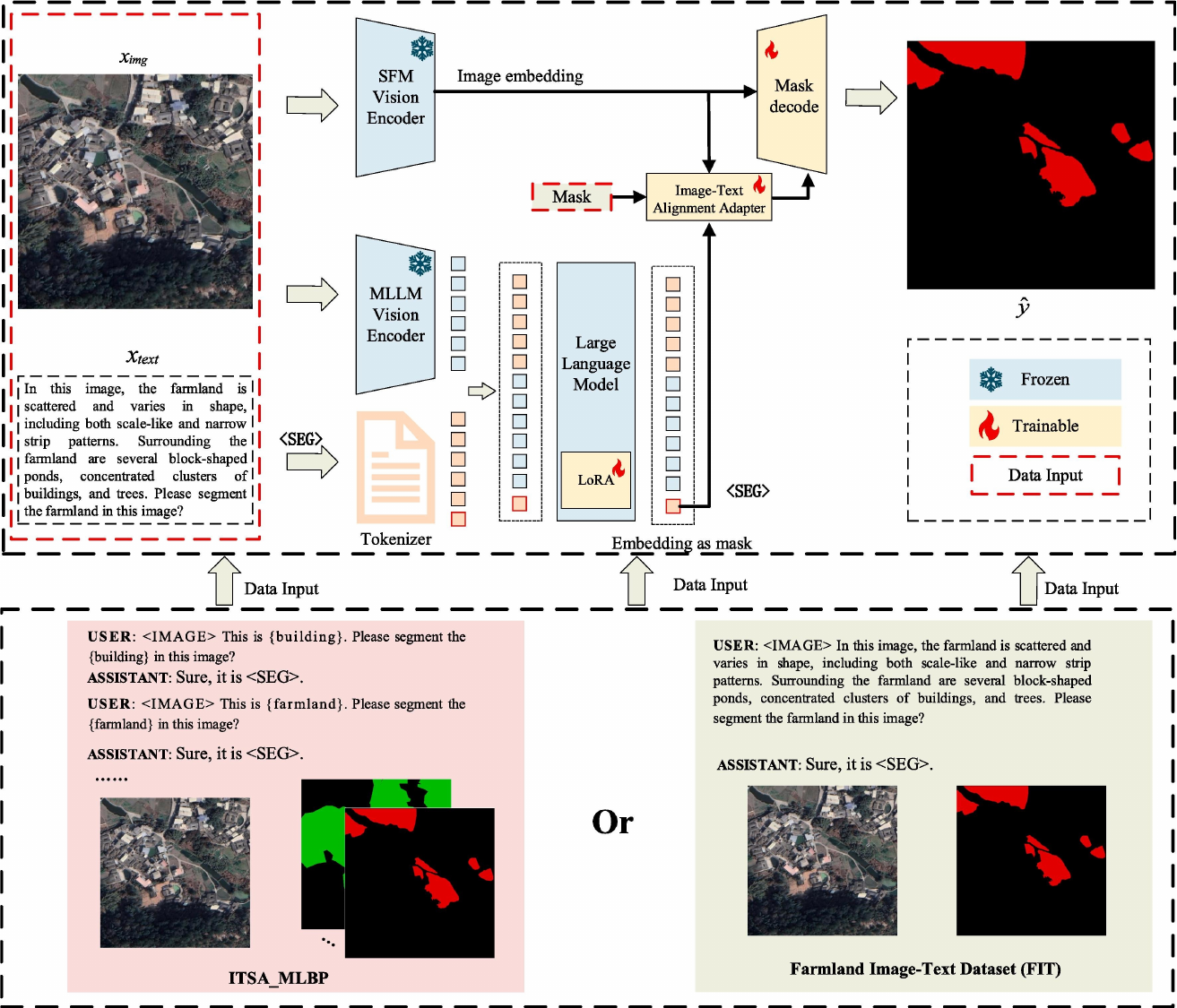}
    \caption{A typical example of a Transformer-based APBD method named FarmSeg$\_$VLM proposed by Wu \textit{et al.} \cite{wu2025farmseg_vlm}.}
    \label{fig:dl_tr}
\end{figure}

In recent years, transformer-based deep learning models, such as Vision Transformer (ViT) \cite{dosovitskiy2020image}, Swin Transformer \cite{liu2021swin}, SegFormer \cite{xie2021segformer} and Relationformer \cite{shit2022relationformer}, have gained attention for their ability in processing and understanding numerous remote sensing data. They leverage multi-head self-attention, which computes pairwise interactions between all patches. Multi-head self-attention makes them able to capture long-range dependencies between geographical features to be suitable for large-scale landscape analysis. Additionally, transformer-based models show promise in few-shot learning, where pretraining and finetuning can mitigate the challenge of limited labeled agricultural parcel and boundary data. 
Some researches use Segment Anything Model (SAM) \cite{kirillov2023segment}, a state-of-the-art image segmentation foundation model, as an auxiliary way to extract features and then adopt traditional machine learning-based methods (such as RFs, SVM, etc) and deep learning-based methods (such as CNN, LSTM, etc.) to significantly improved the performance of classifiers and alleviated the sample scarcity problem \cite{sun2024enhancing}. Ferreira \textit{et al.} \cite{ferreira2025fieldseg} propose a new SAM-assisted crop field extraction framework using 2022 Sentinel-2 temporal composites and presents the lessons learned using this foundational model in eight agricultural regions across the world. The large-scale applicability of this method is demonstrated in four countries (1 million square kilometers), showing promising results and the ability to generalize across different regions. 
Furthermore, in recent years, technological advancements have shown the complementary advantages of vision and language in learning and understanding world knowledge, aiming to integrate the intuitive perception of vision with the deep understanding capabilities of language. Wu et al. design Vision-Language Models (VLMs) for farmland segmentation that combines a semantic segmentation model with a multi-modal Large Language Model (LLM) \cite{wu2025fsvlm}. For example, FarmSeg\_VLM \cite{wu2025farmseg_vlm} designs an image-text spatial alignment strategy under multi-label background priors and an image-text alignment adapter to further correct the spatial alignment mapping between the language descriptions and the corresponding visual features of ground objects (see details in Fig. \ref{fig:dl_tr}. Comparative experiments demonstrated that with the help of language description, the proposed method outperforms existing farmland segmentation methods in both generalization and segmentation accuracy.

In addition, some studies combine different kinds of deep learning-based methods to achieve good connectivity of repair fragmented edges that may appear in semantic edge detection. For example, Xia \textit{et al.} \cite{xia2024crop} both utilize D-LinkNet and Relationformer, which belong to semantic segmentation-based and transformer-based methods, respectively. This method relies on good connectivity to repair fragmented edges that may appear in semantic edge detection with effectiveness and robustness. Commonly, the agricultural field results from traditional machine learning-based or deep learning-based methods are vector surface data; therefore, it is necessary to convert the surface fields into vector contour boundaries. If we directly generated contour edge lines using some software packages (such as QGIS, ArcGIS, etc.), It is not guaranteed to be smooth and continuous and the occurrence of adjacent field boundary adhesion issues. It is critical to generate vectorized agricultural field parcels with closed boundaries from post-processing methods for these models \cite{cai2023improving}. 
For example, Song \textit{et al.} \cite{song2023hierarchical} use Douglas–Peucker Algorithm (DPA) \cite{ebisch2002correction} to optimize the edge contour and generate efficient and stable reconstruction rules of smooth contour lines. DPA is an algorithm that approximates a curve as a series of points and reduces the number of points. In addition, Wang \textit{et al.} \cite{wang2022unlocking} points out that the deep learning model outputs field boundary detections but not separate crop fields. To obtain individual agricultural parcel (also called "instances"), they use a hierarchical watershed segmentation algorithm, which is a region-based algorithm that operates on a grayscale image and treats it like a topographic map, with each pixel’s brightness value representing its height (see details in \ref{sec:tip_region}). 

\section{Assessment}
\label{sec:ass}

\subsection{Public dataset for APBD}

\begin{table*}[t]
  \centering
  \caption{Statistics of representative APBD-related dataset. CI, BD and PS denote three different levels of APBD task, i.e., Cropland Identification, Boundary Delineation and Parcel Segmentation. }
  \label{tab:dataset}

    \begin{tabular}{c c c c c c c c c c}
      %\toprule
      \hline
      \multirow{2}{*}{Dataset} & \multirow{2}{*}{Source} & %\multirow{2}{*}{Year} &
        \multirow{2}{*}{Images} & %\multirow{2}{*}{Image width} &
        \multirow{2}{*}{Area (km$^{2}$)} & {Resolution} &
        \multirow{2}{*}{Classes} & {Global} &
        \multicolumn{3}{c}{Task} \\ %\cline{10-13}
       & & & &  (m) & & coverage & CI & BD & PS  \\ \hline
      %\midrule
      %-------------- Raster mask group ---------------------------------
      %\multicolumn{1}{c|}{\multirow{8}{*}{\rotatebox{90}{ Raster mask}}}
     FHAPD\ \cite{zhao2025large}               & GaoFen-1/2                    & 68,982         & $<$500 & 0.5-1     & 2   &        & $\checkmark$ & $\checkmark$ & $\checkmark$          \\ %& 26522–39842
       GTM\ \cite{li202510}               & Sentinel-2                    & 108,300         & 853,161 & 10     & 2   & $\checkmark$       & $\checkmark$ &  &         \\
     FTW\ \cite{kerner2025fields}                  & Sentinel-2                    & 70,462  &  166,293   & 10      & 2   & $\checkmark$ & $\checkmark$ & $\checkmark$ &  $\checkmark$          \\
      AI4B (S-2)\ \cite{d2023ai4boundaries}  & Sentinel-2                     & 7,831   & 51,321    & 10      & 2   &        & $\checkmark$ & $\checkmark$ & $\checkmark$   \\
      AI4B (Ortho)\ \cite{d2023ai4boundaries}& Aerial                         & 7,598   &  1,992     & 1       & 2   &        & $\checkmark$ & $\checkmark$ & $\checkmark$    \\
      PASTIS\ \cite{hall2024glocab}               & Sentinel-2                    & 2,433      & 3,986     & 10      & 19  &        & $\checkmark$ & $\checkmark$ & $\checkmark$         \\
      CP-Set\ \cite{zhai4940114deeply}               & GF-2                       & 949         & 249        & 1       & 2   &        & $\checkmark$ & $\checkmark$ & $\checkmark$   \\
      GFSAD30\ \cite{thenkabail2021global}              & Landsat-8                     & 64,800      & 18,740,000 & 30     & 2   & $\checkmark$ & $\checkmark$ &        &          \\ 
       GloCAB\ \cite{hall2024glocab}              & Sentinel-2                    &  190,832   & N/A    & 10      & 2   &        & $\checkmark$  & $\checkmark$ & $\checkmark$     \\
      EuroCropsML\ \cite{reuss2025eurocropsml}          & Sentinel-2                     & 706,683    & >4,000     & 10      & 176 &        & $\checkmark$ & $\checkmark$ &             \\
      AI4SmallFarms\ \cite{d2023ai4boundaries}        & Sentinel-2                    & 62       & 1,550     & 10      & 2   &        &  $\checkmark$         & $\checkmark$ &  $\checkmark$           \\
      India Smallholder Boundaries \cite{wang2022unlocking} & Airbus SPOT-6/7   & {10,000}  & {30-50} & {1.5-4.8} & {2} &        &   {$\checkmark$}     & {$\checkmark$} &  {$\checkmark$}        \\
      {{GTPBD \cite{zhang2025gtpbd}}} & {GF-2 \& Google Earth} &   {{47,537}} &  {{885}} & {{0.5–0.7}} & {{3}} & {{$\checkmark$}} & {{$\checkmark$}} & {{$\checkmark$}} & {{$\checkmark$}}  \\
      CropLayer\ \cite{jiang2025croplayer}          & Mapbox \& Google satellite  & 389,777  &9,600,000   & 2    & 2     &     & $\checkmark$\\
    \hline
      %\bottomrule
    \end{tabular}

\end{table*}

\subsubsection{Dataset Construction}
%这里请熠斌帮忙补充
In the construction of a useful APBD dataset, four crucial indexes must be considered. Firstly, the data source serves as the foundation for developing high-quality datasets. As evidenced in selected APBD-related dataset in Table \ref{tab:dataset}, these sources comprise high-resolution Earth observation satellites (e.g., GaoFen, Sentinel, Landsat and Airbus SPOT), aerial images or geographic information programs. Secondly, resolution plays a vital role. APBD tasks necessitate high-resolution images which provide fine-grained boundary information, and most of publications in Table \ref{tab:dataset} have achieved resolutions finer than 30 meter, with the highest resolution reaching 0.5 meters. The third index pertains to the quantity of images. The mentioned datasets in Table \ref{tab:dataset} consistently involve manual annotation of a sufficient number of target samples for training, validation and testing, taking into account the application domain and the complexity of model parameters. Especially for deep learning models which exceed 100 million parameters, such as transformers, an increase in the baseline number of annotated images is advisable. Lastly, the geographic area covered by images is also an important consideration, as it relates to the diversity of agricultural parcel and boundary. Using globally covered images for model training can enhance the model's generalization ability.

\subsubsection{Existing Datasets}
%这里请熠斌帮忙补充
Table \ref{tab:dataset} summarizes key statistics of representative APBD-related datasets, showcasing their diversity in terms of sources, resolutions, and scales. For instance, FHAPD \cite{zhao2025large}, the first large-scale very high-resolution agricultural parcel dataset sourced from GaoFen-1/2, contains 68,982 images covering an area of less than 500 km² with resolutions ranging from 0.5 to 1 m and two labeled classes. On the other hand, datasets like EuroCropsML \cite{reuss2025eurocropsml}, derived from Sentinel-2, stand out with a large number of images (706,683) and 176 labeled classes, emphasizing its versatility. Notably, GTPBD \cite{zhang2025gtpbd}, the first global fine-grained dataset covering major worldwide terraced regions sourced from GaoFen-2 and Google Earth, consists of 47,537 high-resolution images with three-level labels and more than 200,000 complex terraced parcels with manual annotation. The wide range of resolutions and geographic coverage highlights the heterogeneity of APBD datasets. Furthermore, while many datasets are suitable for three different levels of APBD task, it is evident that the variety in parcel types, geographic locations, and image sources leads researchers to develop tailored datasets for specific scenarios. This reinforces the need for standardization and encourages sharing annotated datasets to foster broader adoption.

\subsection{Evaluation on APBD results}\label{sec:evaluation}
%志伟把评价指标的部分改一下，尽量不要跟之前那篇太像，句子表述微调一下

\subsubsection{Pixel-level evaluation metrics}
To evaluate segmentation accuracy on the Crop Identification (\textbf{CI}), five standard pixel-level metrics are commonly used in semantic segmentation tasks, including Precision \textbf{(Prec.)} \cite{chen2024multi}, \cite{li2024comprehensive}, Recall \textbf{(Rec.)} \cite{chen2024novel}, \cite{liu2025lightweight}, Intersection over Union \textbf{(IoU)}\cite{sun2025sidest}, \cite{zhang2025toward}, Overall Accuracy \textbf{(OA)} \cite{li2023using}, \cite{wu2025sbdnet} and \textbf{F1-score} \cite{xie2025integrating}, \cite{zhang2024matnet}.

\textbf{Precision (Prec.)} measures the proportion of correctly predicted agricultural pixels among all pixels predicted as agricultural. A higher precision indicates fewer false positives in land parcel classification.
\begin{equation}
    Prec. = \frac{TP}{TP + FP} \quad 
\end{equation}

where TP and FP indicate true positive and false positive, respectively. TP indicates the number of pixels correctly identified as agricultural parcels, while FP indicate the number of pixels mis-identified as agricultural parcels (i.e., mistakes).

\textbf{Recall (Rec.)} captures the proportion of true agricultural pixels that are correctly identified. High recall reflects a model’s ability to minimize missed detections.
\begin{equation}
    Rec. = \frac{TP}{TP + FN} \quad 
\end{equation}

where FN indicates false negative and the number of pixels mis-identified as non-agricultural parcels (i.e., omissions).

\textbf{Intersection over Union (IoU)} quantifies the spatial agreement between the predicted and ground-truth regions. IoU is widely adopted in segmentation benchmarks due to its robustness to class imbalance.
\begin{equation}
    IoU = \frac{TP}{TP + FP + FN} \quad 
\end{equation}

\textbf{Overall Accuracy (OA)} calculates the proportion of correctly classified pixels across the entire image. OA provides a global view of model performance and is especially informative for datasets with class imbalance.
\begin{equation}
    OA = \frac{TP + TN}{TP + FP + FN + TN} \quad
\end{equation}

where TN indicates true negative and the number of pixels correctly identified as non-agricultural parcels.
  
\textbf{F1-score} is the harmonic mean of precision and recall, offering a balanced evaluation metric particularly useful under domain shift conditions or in presence of noise and uncertainty.

\begin{equation}
    F1 = \frac{2 \cdot Prec. \times Rec.}{Prec. + Rec.} 
\end{equation}

\subsubsection{Edge detection evaluation metrics}

For Boundary Detection (\textbf{BD}) tasks on the APBD scenario, we evaluate model performance using three widely adopted metrics: Optimal Dataset Scale F1-score (ODS) \cite{yan2014automatic}, Optimal Image Scale F1-score (OIS) \cite{yan2014automatic}, and Average Precision (AP) \cite{xu2024multiscale}. These metrics jointly assess the accuracy, adaptability, and robustness of predicted boundaries.

Let $P_t$ and $R_t$ denote precision and recall computed at threshold $t$, and let $F_t$ be the corresponding F1-score:

\begin{equation}
F_t = \frac{2 \cdot P_t \cdot R_t}{P_t + R_t}.
\end{equation}

\textbf{Optimal Dataset Scale F1-score (ODS)} evaluates the global performance of an edge detector across the entire dataset using a single optimal threshold $t^*$:

\begin{equation}
\text{ODS} = \max_{t \in \mathcal{T}} \left( \frac{2 \cdot P_t^{\text{dataset}} \cdot R_t^{\text{dataset}}}{P_t^{\text{dataset}} + R_t^{\text{dataset}}} \right),
\end{equation}

where $P_t^{\text{dataset}}$ and $R_t^{\text{dataset}}$ are aggregated precision and recall over the full dataset under threshold $t$.

\textbf{Optimal Image Scale F1-score (OIS)} computes the mean of the per-image best F1-scores, reflecting local threshold adaptiveness:

\begin{equation}
\text{OIS} = \frac{1}{N} \sum_{i=1}^{N} \max_{t \in \mathcal{T}} \left( \frac{2 \cdot P_t^{(i)} \cdot R_t^{(i)}}{P_t^{(i)} + R_t^{(i)}} \right),
\end{equation}

where $P_t^{(i)}$ and $R_t^{(i)}$ denote the precision and recall on the $i$-th image under threshold $t$, and $N$ is the total number of images.

\textbf{Average Precision (AP)} is computed as the area under the precision–recall curve:

\begin{equation}
\text{AP} = \int_0^1 P(R) , dR,
\end{equation}

where $P(R)$ is the precision as a function of recall, evaluated across all thresholds.

\subsubsection{Object-level geometric metrics}

To comprehensively evaluate the geometric quality of delineated agricultural parcels, we adopt three object-level geometric metrics to evaluate Parcel Segmentation (\textbf{PS}): Global Over-Classification Error (GOC) \cite{lu2024refined}, \cite{li2023using}, Global Under-Classification Error (GUC) \cite{lu2024refined}, \cite{long2024integrating}, and Global Total Classification Error (GTC) \cite{tian2024fieldseg}, \cite{zhao2024irregular}. These indicators quantify segmentation accuracy in terms of spatial overreach, omission, and overall geometric consistency.

Let $S_i$ denote the $i$-th predicted parcel (segmentation), and let $O_i$ represent the ground truth parcel that has the largest intersection area with $S_i$. Denote $m$ as the number of predicted parcels. The object-wise evaluation is defined as follows:

\textbf{Global Over-Classification Error (GOC)} measures the average extent to which predicted parcels exceed the spatial extent of their matched ground truth objects:

\begin{equation}
\text{OC}(S_i) = 1 - \frac{\text{area}(S_i \cap O_i)}{\text{area}(O_i)},
\end{equation}

\begin{equation}
\text{GOC} = \sum_{i=1}^{m} \left( \text{OC}(S_i) \cdot \frac{\text{area}(S_i)}{\sum_{k=1}^{m} \text{area}(S_k)} \right),
\end{equation}

where $\text{area}(\cdot)$ denotes the number of pixels in the respective region.

\textbf{Global Under-Classification Error (GUC)} quantifies the proportion of each predicted parcel not covered by the corresponding ground truth:

\begin{equation}
\text{UC}(S_i) = 1 - \frac{\text{area}(S_i \cap O_i)}{\text{area}(S_i)},
\end{equation}

\begin{equation}
\text{GUC} = \sum_{i=1}^{m} \left( \text{UC}(S_i) \cdot \frac{\text{area}(S_i)}{\sum_{k=1}^{m} \text{area}(S_k)} \right).
\end{equation}

\textbf{Global Total Classification Error (GTC)} synthesizes both over- and under-classification errors into one holistic metric using a root-mean-square formulation:

\begin{equation}
\text{TC}(S_i) = \sqrt{ \frac{ \text{OC}(S_i)^2 + \text{UC}(S_i)^2 }{2} },
\end{equation}

\begin{equation}
\text{GTC} = \sum_{i=1}^{m} \left( \text{TC}(S_i) \cdot \frac{\text{area}(S_i)}{\sum_{k=1}^{m} \text{area}(S_k)} \right).
\end{equation}

\section{Discussions}
\label{sec:diss}

APBD is of utmost importance for a comprehensive understanding of the food security on both global and local scales. The meta-analysis presented is convenient to outline the past, current and potential future of APBD for those who want to know about this specific domain. A thorough introduction of APBD algorithms in this review may be interesting to them. In this section, we discuss four APBD-related issues to further comprehend the APBD domain.

\begin{table*}[t]
    \centering
    \caption{Qualitative comparison among different APBD methods in annotation, efficiency and accuracy. \textbf{+++} denotes the method performs best in this respect, while \textbf{+} denotes the method performs worst in this respect.}
    %\resizebox{\textwidth}{!}{
    \begin{tabular}{ccccc}
    \hline
        \multicolumn{2}{c}{Method}  & \multirow{1}*{Annotation} & \multirow{1}*{Efficiency} & \multirow{1}*{Accuracy}  \\
        %Class & Division & Section \\
        \hline
       \multirow{2}*{Traditional image processing-}  & Pixel-based & +++ & +++ & +  \\
       \multirow{2}*{based APBD methods}& Edge-based & +++ & +++ & + \\
       & Region-based &  +++ & ++ & ++ \\ %\hline
       \multicolumn{2}{c}{Traditional machine learning-based APBD methods} & ++ & ++ & ++\\ %\hline
       \multirow{2}*{Deep learning-} & Semantic segmentation-based &  + & + & +++\\
       \multirow{2}*{based APBD methods}& Object detection-based & ++ & + & +++ \\
       & Transformer-based & + & + & +++ \\
       \hline
    \end{tabular}
    %}
    \label{tab:comparison}
\end{table*}

\subsection{Multi-sensor data in APBD domain}
\label{sec:multi}

Due to the extreme imbalance of categories in farmland boundary extraction tasks, the direct use of neural networks for boundary pixel prediction in remote sensing images often results in poor performance. Therefore, some studies investigate the beneficial effects of multi-source and multi-sensor remote sensing data fusion to assist in boundary extraction. 
In addition, most of the existing research on farmland boundary extraction is based on optical remote sensing data. The quality of optical remote sensing images depends on the conditions of no cloud or rain \cite{liu2020farmland}. Research in rainy and cloudy areas will be limited to some extent. Compared with this, remote sensing platforms based on polarized SAR are not limited to cloud and rain conditions, and have the characteristics of all weather conditions. 
For example, Kuang \textit{et al.} \cite{kuang2023agricultural} propose a novel cascaded model of semantic segmentation and edge detection for the boundary extraction task of farmland in northern Xinjiang, using the fused data of Sentinel-1 dual-polarized SAR images with multiple temporal phases and Sentinel-2 multi-spectral images.
In addition, some studies also utilize different optical remote sensing images with multi-resolution to extract different levels of features for cropland and boundary. Very High Resolution (VHR) images (such as QuickBird, WorldView, etc.) could easily distinguish field boundaries in areas with established croplands, while processing large volumes of VHR scenes for a region is computationally intensive, and has been previously a cost prohibitive task. One the other hand, Although moderate resolution satellite data (such as Landsat, Sentinel, etc.) do not have adequate spatial resolution to capture the total area occupied by smallholder farms, they have richer spectral information and time-series temporal characteristics with easy accessibility. 
For example, Mccarty \textit{et al.} \cite{mccarty2017extracting} combine the results from sub-meter WorldView-1 and WorldView-2 segmentation with median phenology amplitude from Landsat 8 data to map cropped area for the rainfed residential cropland mosaic of the Tigray Region, Ethiopia, which is comprised mostly of smallholder farms.

\subsection{Single-task learning v.s. Multi-task learning}

Currently, deep neural networks have been widely used in extracting thematic information due to their powerful feature extraction capabilities \cite{long2022delineation,sumesh2025novel}, which could be categorized into two classes: i.e., single task and multitask models. Single-task networks typically use an encoder-decoder architecture, where the encoder captures multi-scale semantic features, and the decoder subsequently refines these gathered features \cite{zhang2020generalized,zhang2021automated}. Although these approaches produce satisfactory results, they tend to overlook the benefits of intrinsically related tasks that share optimal hypothesis classes, which could enhance generalization and reduce overfitting risk.
Therefore, researchers have increasingly adopted multi-task architectures that allow for joint representation learning across multiple related tasks \cite{waldner2020deep,li2023using,pan2024rbp}. For example, Long et al. \cite{long2022delineation} propose BsiNet that learns three tasks: a core task for agricultural field identification and two auxiliary tasks for field boundary prediction and distance estimation, corresponding to mask, boundary, and distance tasks, respectively. However, such method fails to fully consider the multi-level boundary semantics, resulting in the incomplete and unclosed parcel boundaries. To deal with these deficiencies, the dual-branch architecture perform the tasks of agricultural parcel detection and boundary delineation separately, using a shared encoder but distinct decoders, which improves the representation of
boundary features, thus enhancing the accuracy of AP delineation in diverse areas (such as SEANet \cite{li2023using} and HBGNet \cite{zhao2025large}).

\subsection{Comparison among different APBD methods}

Table \ref{tab:comparison} displays a qualitative assessment for different APBD methods in three aspects: annotations, efficiency, and accuracy. Here we conduct in-depth discussions on them.

\subsubsection{Annotation}

It is necessary and fundamental to conduct the annotation work in supervised learning. Traditional image processing based APBD methods have the least cost, and most of them are unsupervised learning methods and do not require any annotation work \cite{ji1996delineating,yan2016conterminous}. Annotation work of deep learning-based APBD methods (especially for semantic segmentation-based and Transformer-based methods) is the most difficult and complex among all methods, since most of them are pixel-level annotation and have to carefully outline all kinds of fine-grained boundary shapes for agricultural parcels \cite{zhang2020generalized,jong2022improving}. As for most traditional machine learning-based APBD methods, we not only have to annotate the samples of farmland, but also have to annotate the samples of other land cover types, such as bare land, water, trees, impervious area, etc. As for object detection-based methods, we have to annotate the location of the four corners of a agricultural parcel and generate a bounding box for each individual parcel \cite{wang2024aams}. Of course for Mask R-CNN and Mask2Former, we further have to annotate the thorough shape of each agricultural parcel to conduct parcel segmentation \cite{zhong2023multi,cao2023case}. Above two mentioned APBD methods are of moderate difficulty and their annotation works are more difficult than traditional image processing-based APBD methods while easier than semantic segmentation-based and Transformer-based APBD methods.

\subsubsection{Efficiency} 

Algorithm efficiency is a crucial and key factor in APBD applications, especially applied to large-scale study area. Since most of them are unsupervised learning algorithms, traditional image processing-based APBD methods cost most of time in simple and basic image operations, usually with low computation complexity \cite{watkins2019comparison}. However, region-based methods may cost more iteration times than pixel-based and edge-based methods. Therefore, the efficiency of region-based methods is lower than other two traditional image processing-based methods. 
Transformer-based APBD methods have worst performance on algorithm and implementation efficiency, given that their parameters are huge, although their performance is the best. 
Traditional machine learning-based APBD methods usually require the time-consuming sliding window scheme to achieve the location and recognition of cropland \cite{teluguntla201830}. In addition, classifiers or neural networks training  and parameter tuning phases worsen their efficiency. Although semantic segmentation-based and object detection-based APBD methods both have time-consuming neural networks training work, they belong to end-to-end algorithms that allow the delineation of agricultural boundaries in the whole patch image \cite{zhao2024irregular}. To this end, these three algorithms are moderately efficient, higher than Transformer-based APBD methods while lower than traditional image processing-based APBD methods.

\subsubsection{Accuracy} 

APBD accuracy is the most important evaluation to judge the APBD algorithm whether successfully applied to practical agricultural inventory. In general, deep learning-based methods perform best in accuracy, with high capacity of robustness and generalization \cite{liu2022deep,lu2024refined}. Specially, deep learning-based APBD methods achieve more convincing and satisfying results in complex area \cite{persello2019delineation,pan2023e2evap}. Notably, Transformer-based APBD method hold a litter better performance than semantic segmentation-based and object detection-based APBD methods. As for traditional image processing-based APBD methods, their accuracy is generally the lowest among different algorithms, and only have satisfied performance in simple area or under specific parameters or specific regions. When the study sites turn to a varied topography, complicated environment or regions with overlapping crowns, the accuracy may incur a terrible deterioration. It is noteworthy that region-based methods usually perform best compared to pixel-based and edge-based methods \cite{wang2023survey}. The accuracy of traditional machine learning-based APBD methods is between that of deep learning-based APBD methods and traditional image processing-based APBD methods.

\subsection{Comparisons between general deep learning models and their applications in APBD domain}

\begin{figure*}[t]
    \centering
    \includegraphics[width=1.0\linewidth]{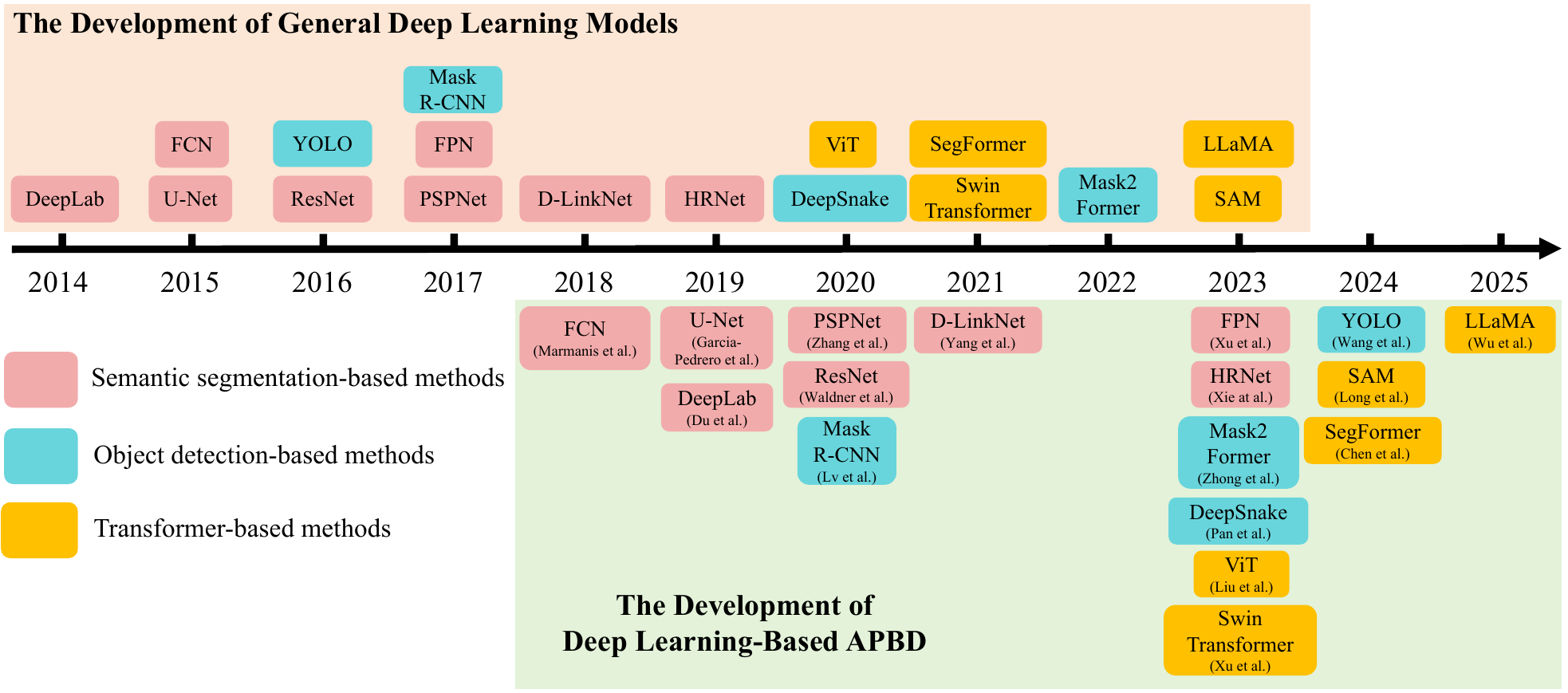}
    \caption{Comparisons of general deep learning models and their applications in the APBD domain.}
    \label{fig: different deep learning methods in APBD}
\end{figure*}

In Figure \ref{fig: different deep learning methods in APBD}, the top of the timeline shows the development of general deep learning architectures, and the bottom shows the years that these deep learning models were first used in the APBD domain. 

Early research in APBD was predominantly driven by semantic segmentation-based methods. These approaches frame the task as a pixel-wise classification problem, aiming to assign each pixel to a specific class, such as 'parcel' or 'background'. As depicted, foundational models like FCN, U-Net, and DeepLab were among the first to be adapted for this purpose between 2018 and 2020. Subsequently, more sophisticated architectures such as PSPNet, D-LinkNet, and HRNet, which offer improved context aggregation and multi-scale feature representation, were also introduced to enhance the accuracy and completeness of the extracted parcel boundaries.

A subsequent evolution in methodology saw the adoption of object detection-based frameworks, which treat each agricultural parcel as a distinct 'instance'. This paradigm shift enables the delineation of individual, non-overlapping field boundaries, a critical requirement for parcel-level analysis. Pioneering instance segmentation models like Mask R-CNN were first applied to APBD around 2020. This was later followed by more advanced contour-based methods such as DeepSnake, which directly regresses the boundary vertices, and sophisticated mask-prediction models like Mask2Former and YOLO, further improving the precision of individual parcel extraction.

The most recent phase in APBD research is characterized by the integration of Transformer-based architectures. These models, leveraging self-attention mechanisms, excel at capturing long-range dependencies and global contextual information, which is highly beneficial for understanding the complex layouts of agricultural landscapes. Furthermore, the advent of large foundation models has introduced novel paradigms. The Segment Anything Model (SAM) offers powerful segmentation capabilities through prompting. Similarly, Large Language Models (LLMs) like LLaMA represent a significant shift. In the context of APBD, LLaMA is not directly used for pixel-level segmentation; instead, it is typically fine-tuned using techniques like LoRA to function as an advanced instruction-following component. It interprets a user's textual requirements—such as "extract only the irrigated rice paddies"—and guides an underlying visual segmentation model to perform the specific extraction task, thereby enabling a more interactive and semantically-aware parcel delineation process.

According to different deep learning models, we can complete different APBD tasks (see Table \ref{tab:comparison}). General deep learning models can be directly applied to APBD scenarios. However, the following differences and modifications should be considered:

\begin{enumerate}
    \item Since the size of a remote sensing image is too large to be data input for general deep learning model, we need to utilize an overlapping partition method for a large-scale remote sensing image into several sub-images in inference phase. After that, we apply coordinates transformation and merge results of all sub-images to achieve the final APBD results \cite{zhao2025large,wang2022unlocking}.
    \item As for the design of deep learning architectures, we need to modify the sizes and the ratios of candidate anchor boxes in object detection-based APBD methods, due to the size of agricultral parcel is usually different from general objects. Furthermore, the number of channel in the first layer usually need to be modified because of multi-spectral bands for remote sensing images rather than only three bands for general images \cite{masoud2019delineation,ferreira2025fieldseg}.
    \item Many deep learning-based need to have some post-processing procedures. For example, the results of semantic segmentation-based APBD methods is a “confidence map”, so that they usually require the DPA \cite{song2023hierarchical} or watershed segmentation \cite{wang2022unlocking} to produce the final detection and delineation of individual parcels. Some studies design a specific post-processing vector generation and homogeneity checking  form closed and discrete polygons \cite{zhang2025toward} and improve the accuracy of APBD model.
\end{enumerate}

\subsection{Criteria for Choosing Appropriate APBD Methods} \label{sec:factors}

Due to the complexity of different research subjects (e.g., mixed cropland, specific type of cropland, and so forth) with different attributes (e.g., areas, density, locations, and so on), it is desirable to design or select appropriate APBD approaches that address APBD tasks under different scenarios. In this section, we discuss multiple influencing factors of APBD approaches.

\subsubsection{Crop types}

The accuracy of APBD is significantly influenced by the type of crops grown within each field. Different crops exhibit distinct canopy structures, texture features, and planting patterns, which directly affect the separability of parcel boundaries in remote sensing imagery. For example, large-scale crops such as maize and wheat are typically grown in regular rows with uniform canopy coverage, resulting in clear boundaries and homogeneous textures that facilitate accurate segmentation. In contrast, crops like vegetables, legumes, and mixed grains are often grown in small fields with diverse management practices such as inter-cropping, crop rotation, or multiple harvests per year. These conditions lead to spectral heterogeneity and ambiguous boundaries, thereby reducing segmentation performance. 
The phenological stage of crops also plays a crucial role. For instance, paddy fields exhibit distinct boundaries during the early transplanting phase due to exposed soil and water backgrounds, but these boundaries become less distinguishable in later stages when the canopy closes and inter-parcel differences diminish. %Moreover, field management practices affect boundary clarity: large, mechanized farms with single-crop cultivation tend to produce regular, well-defined parcels, whereas smallholder systems with fragmented and irregular plots pose additional challenges for model generalization.
%To address these issues, %edge-aware techniques such as boundary loss functions and conditional random field (CRF) post-processing can be employed to enhance model sensitivity to boundary details. Additionally, 
the fusion of multi-source remote sensing data, particularly combining optical and Synthetic Aperture Radar (SAR) imagery, can improve robustness in complex agricultural scenarios by leveraging complementary spatial and spectral features. Overall, crop type is a key factor influencing parcel extraction performance, and careful consideration of crop-specific characteristics is essential for developing models with high generalizability and accuracy across diverse agricultural landscapes.

\subsubsection{Terrain and topography}

Terrain and topography characteristics play a critical role in the accuracy of APBD from remote sensing imagery. These effects are manifested in the spatial morphology of parcels, the spectral expression of land surfaces, and the variability of cropping patterns across different landscapes. In flat areas, such as the North Plain or the Northeast Plain in China, agricultural fields tend to be large, regularly shaped, and managed uniformly. These conditions produce clear spectral signatures and well-defined boundaries in remote sensing images, enabling high-precision extraction using deep learning models. In contrast, in hilly and mountainous regions, parcels are often small, irregularly shaped, and constructed along slopes. These landscapes result in fragmented spatial patterns and complex boundary shapes, which significantly increase the difficulty of accurate delineation \cite{zhang2024dual}. 
Topographic variation also affects the radiometric properties of remote sensing imagery. In areas with significant elevation changes, differences in slope and aspect alter the illumination geometry, causing uneven reflectance even within the same crop type. This intra-parcel spectral inconsistency can lead to over-segmentation or misclassification. Additionally, terrain-induced shadows and occlusions are common in mountainous areas, obscuring boundary information and introducing errors in parcel segmentation. These challenges are particularly pronounced when using single-date optical imagery, where shadow and lighting variation are not compensated for, resulting in reduced accuracy and robustness. 
%To address these challenges, integrating auxiliary terrain information has shown promising results. Incorporating digital elevation models (DEMs), slope, and aspect maps as additional input channels can enhance spatial awareness and boundary sensitivity, particularly in topographically complex areas. 

\subsubsection{Cropping practice and management}

Cropping practice and management also exert a significant influence on the performance of APBD by shaping field characteristics such as size, shape, and boundary clarity. In regions dominated by mechanized farming, such as large-scale grain-producing areas, agricultural parcels tend to be large, geometrically regular, and uniformly managed. These fields often feature well-maintained physical boundaries such as irrigation canals, roads, and embankments, which translate into distinct and stable edges in remote sensing data. As a result, spectral signatures within these parcels are spatially homogeneous and texturally consistent, facilitating accurate and reliable delineation by segmentation models. 
%The uniformity of planting schedules and crop growth stages in mechanized systems further contributes to reduced spectral variability, enhancing the ability of models to identify clear parcel boundaries. 
Conversely, smallholder farming systems, commonly found in mountainous or hilly regions and parts of developing countries, present markedly different challenges. Here, parcels tend to be smaller, irregularly shaped, and fragmented, reflecting diverse land ownership and heterogeneous management \cite{zhao2024irregular}. In such systems, fields may undergo mixed cropping, multiple planting cycles, leading to complex within-field spectral heterogeneity. Moreover, the absence or poor maintenance of physical boundaries such as embankments and irrigation infrastructure results in ambiguous and often discontinuous edges in the imagery. These factors contribute to higher intra-parcel variability and diminished boundary clarity, which substantially reduce the accuracy of automated parcel extraction. 
Furthermore, regional agricultural traditions and practices impact the spatial configuration and management of parcels. For example, in Southeast Asian upland areas, fragmented landholdings and mixed-cropping are prevalent, resulting in complex parcel patterns that challenge remote sensing-based extraction. In contrast, regions characterized by industrial-scale monoculture farming typically exhibit larger, more uniform fields, which are more amenable to remote sensing delineation. This regional variability implies that models trained on data from mechanized, large-scale farming regions may not generalize well to smallholder or mixed management systems without adaptation. 
%To improve model robustness across diverse agricultural landscapes, incorporating information about local management practices and boundary conditions is essential. Strategies such as integrating ancillary data on field boundaries, utilizing multi-temporal imagery to capture phenological variation, and applying adaptive segmentation algorithms can help overcome challenges posed by heterogeneous management systems. 
%The condition and maintenance of parcel boundaries themselves also play a crucial role. Well-maintained physical boundaries improve edge detectability in remote sensing data by providing sharp spatial transitions that segmentation algorithms can exploit. Conversely, in regions where boundaries are poorly maintained or absent, natural vegetation growth, erosion, or human activity may blur or eliminate these features, increasing boundary ambiguity and segmentation errors.

\section{APBD related Applications}
\label{sec:application}

In this section, we introduce some practical APBD-related applications. Other APBD applications include wildfire potential estimation, plant heterogeneity, wildlife protection and biodiversity research, and so forth. Most of them first conduct agricultural parcel and boundary detection and then further conduct other analyses on a single-plot scale. With the recent emergence of end to-end deep learning techniques, we are able to achieve APBD, along with cropland classification or yield estimation in the meantime.

\subsection{Crop type classification}

APBD task is a preliminary step in crop type classification, which holds significant implications for cadastral management and agricultural land governance \cite{crommelinck2017contour,wassie2018procedure}. However, most of existing crop type classification is based on the first level of APBD, crop identification, without any boundary delineation or parcel extraction. In addition, existing parcel-level crop type classification is a two-stage framework, where they firstly conduct APBD for regions \cite{alganci2013parcel,jiao2022parcel} or directly using cropland parcel boundary vector data from others \cite{kussul2016parcel,zhang2020parcel,zhang2021improving}, and then get the crop types for each individual parcel using another classifier or model. For example, Xie \textit{et al.} \cite{xie2025integrating} firstly adopt a boundary-enhanced based on the U-Net model for farmland parcel extraction from Gaofen-2 data, and then select RFs to achieve crop
classification at the parcel level from both Sentinel-2 and Landsat 8 data. Therefore, as deep learning techniques emerged, an end-to-end parcel-level crop type classification would be proposed to apply in larger area with high-efficiency.

\subsection{Yield estimation}

%Crop production is forecast with the support of crop area estimates and yield predictions for specific combinations of agro-ecological regions, administrative units and crop types. Crop type mapping and geo-statistical methods are two categories of methods to derive crop area estimations, while crop type mapping not only provides data to estimate crop area, but also provides baseline data for crop condition assessment and yield prediction.

Accurate and timely crop yield estimation are critical to realizing global food security, balancing international grain trade, and promoting sustainable agricultural development \cite{xiao2025progress}. By providing consistent and large-scale observations, remote sensing technology has become indispensable in crop yield estimation across local, regional, and global scales. Over the past four decades, numerous crop yield estimation approaches have been developed. It is of the greatest potential of integrating artificial intelligence and remote sensing technologies with process-based crop growth models through data assimilation techniques. However, most of existing yield estimation studies are limited on pixel-level instead of parcel-level. Some parcel-level yield estimation only conduct experiments on very small scale area \cite{liu2018mapping,wang2019rice}, and may adopt existing open-source field parcels boundaries dataset \cite{hiremath2021crop} or manually delineation \cite{lambert2017estimate}, which is actually a two-stage scheme. We think it is important to develop some end-to-end parcel-level yield estimation algorithms using advanced deep learning methods in the future.

\subsection{Stress detection and health monitoring}

Multi-spectral information from remote sensing data, coupled with machine learning or deep learning techniques, plays a considerable role in crop stress \cite{berger2022multi,liu2018mapping}, crop phenology \cite{gao2021mapping,d2022monitoring} and other health monitoring \cite{wu2023challenges}, including disease nutrients, and pests surveillance, growing status observation \cite{d2018targeted}, and so forth. Similar to cropland classification, previous popular individual parcel’s health monitoring is two-stage work, although, as deep learning-based APBD methods emerge, existing crop’s health assessment becomes an end-to-end one-stage framework, accomplishing both individual parcel delineation and their status observation. Compared to SAR/Lidar data, deep learning may perform better on multi-spectral optical remote sensing data because of its rich semantic and texture information, which is also beneficial for capturing vegetation’s intrinsic features. As a matter of fact, employing comprehensive health monitoring is beneficial to improve their productivity or yield, and further increase the economic effect.

\subsection{Multi-temporal change detection}

On the one hand, multi-temporal remote sensing images are able to conduct more accurate agricultural parcel extraction from time-series features \cite{yang2021semantic,yan2024tsanet,tang2025parcel}. For example, Garnot \cite{garnot2022multi} propose a multi-modal temporal attention-based model by leveraging both optical and radar time series to outmatch single-modality models for crop mapping in terms of performance and resilience to cloud cover. 
On the other hand, multi-temporal images could also explore the changes of cropland area \cite{potapov2022global}, crop types, growth process \cite{gao2021mapping}, and so forth, evaluating the variants of food production \cite{you2025climate} and carbon stock \cite{beyer2022relocating} or the impacts of abandoned cropland \cite{zheng2023neglected}, natural disasters \cite{xu2021spatiotemporal} and conservation policies\cite{schneider2024effects}. Also, multi-temporal data contribute to better accomplishing APBD and APBD-related applications through seasonal spectral and texture variations \cite{asgarian2016crop,zhong2019deep,wang2023new}. 
However, similar to crop classification, existing single parcel-level change analyses are all two-stage works \cite{sun2024enhancing}. Following the development of recurrent neural networks, we believe that coupling semantic segmentation-based and object detection-based APBD methods with time-series analysis may achieve real-time, high-accuracy, and end-to-end single parcel-level change analysis using multi-temporal remote sensing data.

%\subsection{Associated with climate change}

\section{Prospects}
\label{sec:pros}

Based above literature analysis, methodology review and in-depth discussion, APBD-related prospects emerged from this attempt, which concerns past, current and future trends. These prospects are introduced in this section. 

\subsection{Multi-source and multi-resolution remote sensing data fusion}
\label{sec:msmr}
The integration of multi-source and multi-resolution remote sensing data is expected to play a transformative role in advancing APBD. Traditional approaches often depend on single-source optical imagery, which is constrained by cloud cover, spectral limitations, and inconsistent resolution \cite{liu2020farmland}. In contrast, data fusion enables the combination of complementary information across diverse platforms, including VHR optical imagery, medium-resolution time series satellites (e.g., Sentinel-2, Landsat), SAR, LiDAR, and hyperspectral sensors, providing richer spatial, spectral, and temporal signals to support more accurate and robust delineation \cite{liu2020farmland}. The synergy between spatial detail and temporal continuity improves both intra-field homogeneity and inter-field separability, which are essential for high-quality segmentation. 
On the other hand, recent advances in deep learning facilitate more effective fusion of heterogeneous data \cite{yuan2024fusu}. Transformer-based models and attention mechanisms can learn from multi-resolution and multi-modal inputs, dynamically weighting relevant features according to the landscape context \cite{zhang2024matnet}. This enables more generalizable and adaptive parcel extraction models across diverse agro-ecological zones, especially when applied at continental or global scales. 
Therefore, the fusion of multi-source and multi-resolution remote sensing data represents a critical frontier for the future of APBD. Building intelligent, adaptable, and transferable frameworks will be essential for ensuring accurate delineation across variable landscapes, supporting both scientific research and operational applications in agriculture and land management.

\subsection{Knowledge-guided explainable deep learning-based method for APBD}

As remote sensing technologies evolve toward higher spatial, spectral, and temporal resolutions, the challenge of accurate APBD becomes increasingly complex. In this context, knowledge-guided and explainable deep learning methods are emerging as a promising direction to improve the interpretability, generalizability, and reliability of models. Unlike purely data-driven approaches, knowledge-guided deep learning integrates domain-specific priors, such as agronomic patterns, crop growth cycles, and cadastral logic, into the model design to enhance performance in heterogeneous and data-scarce regions \cite{wu2023multilevel}. In addition, such models not only learn from raw imagery but also from structured knowledge graphs, spatial rules (e.g., field shape regularity, adjacency constraints), and temporal behaviors (e.g., planting-harvest cycles) \cite{garcia2018outlining}. This fusion of learning and reasoning facilitates more accurate detection of subtle or ambiguous field boundaries, especially in smallholder farming systems or terraced areas. Moreover, by embedding explainable components such as attention visualization, saliency maps, and rule-based decision modules \cite{kakogeorgiou2021evaluating}, these methods offer insights into model behavior, supporting trust and transparency in agricultural monitoring applications.
In future developments, knowledge-guided models will be instrumental in enabling scalable, adaptive, and policy-aligned parcel extraction pipelines, and explainability mechanisms will support user validation, integration with cadastral databases, and collaborative decision-making in precision agriculture and land management.

\subsection{Fine-grained crop or growing status classification}

Fine-grained cropland classification includes both fine-grained crop \cite{sun2020geo} and fine-grained growing status (such as different levels of disease, stress and damage, etc.) classifications. 
The former is significant for understanding the distribution of crop types and ensuring food production. The latter not only observes damaged or diseased crops to prevent their proliferation but is also conducive to estimating yield and increasing the benefits for some economic crops. %To this end, fine-grained crop classification is extremely valuable to both ecology and economy. 
%Most of the existing crop classification studies focus on small areas (smaller than 1,000 ha). Although Zhang et al. [185] estimate the number of tree species in tropical areas, they have not mapped the distribution of fine-grained tree species. 
%需要加一段crop classification的表述
As for growing status observation, most of them are classified into only two statuses: healthy and unhealthy croplands. By contrast, few studies are devoted to multiclass growing status classification. It is highly demanded for cropland to monitor more fine-grained healthy conditions, such as specific diseases. Until now, individual agricultural parcel classification work has been a two-stage scheme, first detecting or delineating the agricultural parcel, and then completing types or growing status recognition \cite{d2018targeted}. Furthermore, fine-grained classification requires recognizing the slighted difference between similar classes, which may need high spatial- and high spectral-resolution remote sensing images. The data-fusing approaches mentioned in the Sec. \ref{sec:multi} and Sec. \ref{sec:msmr} would be an effective solution.

\subsection{Large-scale APBD in spatial big data era}

Undoubtedly, we are presently in the big data era and will continue to be so in the future. Massive remote sensing images acquired by satellites, aerial planes, UAVs, and even mobile phones create the spatial big data era. With these Earth-observation data, we have the opportunity to achieve large-scale APBD and deeply comprehend global cropland resources. Although there are some global \cite{fritz2015mapping}, national \cite{tu202430} or large-scale regional \cite{zhang2020generalized} crop mapping products. 
%Most of the study areas in existing APBD research are smaller than 1,000 km$^2$, except for those in \cite{zhang2020generalized}. They extract agricultural parcels over a land area that spans 940,000 km$^2$ in four major provinces in China. 
However, these works only focus on cropland identification instead of fine-grained boundary delineation or individual parcel delineation. 
For example, despite the fact that Potapov \textit{et al.} \cite{potapov2022global} estimate that there are roughly 1,244 Mha around the world in 2019, they only approximately map the global cropland distribution and mapping without fine-grained agricultural parcel delineation. There are two major challenges in large-scale APBD work. 
The first is the capacity of model generalization. As we have to prepare multi-temporal, multi-source, and multi-regional remote sensing data to conduct large-scale APBD, developing a more transferable, robust, and general model is a powerful foundation, using advanced algorithms such as domain generalization \cite{zheng2021multisource,li2024hyunida}, domain adaptation \cite{zheng2024open,li2025boosting} and transfer learning \cite{zheng2022partial,liang2025low}. Also, how to transfer our model to apply in complex areas, such as smallholder farmlands, terrace and mountain areas \cite{zhang2025gtpbd}, is really difficult and data-scarcity. 
Another challenge is the capacity of computation performance to support the efficiency of APBD in large-scale areas. At present, seldom study adopts high-performance computation platforms (such as Field Programmable Gate Arrays (FPGAs) and GPUs) to accelerate APBD algorithms. However, global, continental, or national-level APBD research has not been completed in higher-performance computing platforms such as supercomputers, which may be a potential general platform for processing global observation issues \cite{zheng2023achieving,zhao2023sw}.

\subsection{Foundation models for APBD}

The rise of large-scale foundation models, including Visual Foundation Models (VFMs) \cite{guo2024skysense} and Vision-Language Models (VLMs) \cite{zhang2024vision}, is transforming remote sensing and geospatial analysis. For APBD domain, these models offer unprecedented potential to improve generalizability, reduce reliance on labeled data, and enhance interpretability. Pretrained on massive and diverse datasets, VFMs can capture robust visual representations of spatial patterns and textures that generalize across regions, seasons, and crop types. This makes them well-suited to address the variability inherent in agricultural landscapes, especially in smallholder-dominated or fragmented regions. 
More recently, multi-modal VLMs (such as CLIP, GPT-4o, etc.) have demonstrated the ability to jointly process imagery and textual knowledge, enabling high-level semantic understanding of land features \cite{wu2025farmseg_vlm}. In the context of parcel delineation, VLMs can integrate remote sensing imagery with auxiliary textual inputs such as crop calendars, agronomic rules, or cadastral descriptions to infer semantically meaningful boundaries. This paves the way for more context-aware, flexible, and human-aligned delineation systems. 
Additionally, these large models support in-context learning and zero-shot transfer \cite{liu2024remoteclip}, reducing the need for task-specific training and allowing scalable deployment in new regions. Therefore, foundation models are expected to become more and more important in APBD. When fine-tuned with domain-specific datasets or prompted with expert knowledge \cite{dong2024upetu}, they will enable intelligent, adaptive, and interpretable parcel extraction workflows that are capable of supporting diverse use cases from precision agriculture to large scale land governance.

\section{Conclusions}
\label{sec:concl}

APBD using high-resolution remote sensing data is essential for precise agriculture and agricultural resource management in an automated way. In this review article, a comprehensive overview of APBD-related research was introduced. First, we conducted an investigation of scientific peer-reviewed journal articles over 20 years, building an available database and carrying out a meta-analysis. Second, intriguing and thorough APBD methods that depict the trend and development of past years relating to this specific domain were presented. We classify APBD methods into three types: traditional image processing based (such as watershed algorithm, image segmentation, and so on), traditional machine learning based (such as RF and DT and so forth), and deep learning based. In addition, we also categorized deep learning-based APBD methods into two types (i.e.,
semantic segmentation and object detection) and discussed their pros and cons. At the current pace that the methodology of APBD research is conducted, such information is rather essential and truly valuable. In addition, we discussed three APBD-related topics to further comprehend the APBD domain, such as comparisons between active remote sensing data and passive remote sensing data, comparisons among different algorithms, and different APBD tasks. Finally, some APBD-related applications and a few existing and emerging topics were presented, and we promise the significance and prosperity of APBD in the future.

% or


%\appendices

%\section{}

%\section*{Acknowledgment}

%The authors would like to thank Hui Lu, Xiaomeng Huang and Mengyao Sun for their valuable discussions.

\ifCLASSOPTIONcaptionsoff
  \newpage
\fi




%\begin{thebibliography}{1}

%\small
%\bibliographystyle{plainnat}
\bibliographystyle{IEEEtran}
\bibliography{refs}
\end{document}